\documentclass[accepted]{uai2022} % after acceptance, for a revised
\usepackage[american]{babel}
\usepackage{natbib} % has a nice set of citation styles and commands
\usepackage{mathtools} % amsmath with fixes and additions
\usepackage{amssymb, amsthm}
\usepackage{bm}
\usepackage{multirow}
\usepackage{booktabs} % commands to create good-looking tables
\usepackage[ruled, vlined, linesnumbered]{algorithm2e}
\let\oldnl\nl% Store \nl in \oldnl
\newcommand{\nonl}{\renewcommand{\nl}{\let\nl\oldnl}}% Remove line number for one line

\hypersetup{hidelinks}

\newcommand{\setA}{{\mathcal{A}}}

\newcommand{\doubleE}{{\mathbb{E}}}

\newcommand{\vecG}{{\textbf{G}}}

\newcommand{\setM}{{\mathcal{M}}}

\newcommand{\doubleR}{{\mathbb{R}}}

\newcommand{\setS}{{\mathcal{S}}}

\title{On Credit Assignment in Hierarchical Reinforcement Learning }

\author[1, 2]{\href{mailto:<J.A.deVries@tudelft.nl>?Subject=Multilevel Hierarchical Q(lambda) UAI 2022 paper}{Joery~A.~de~Vries}{}}
\author[2]{Thomas~M.~Moerland}
\author[2]{Aske Plaat}
\affil[1]{%
    Algorithmics Group \\
    TU Delft \\
    Delft, The Netherlands
}
\affil[2]{%
    Leiden Institute of Advanced Computer Science \\ Leiden University\\
    Leiden, The Netherlands
}
  
\begin{document}
\maketitle

\begin{abstract}
 Hierarchical Reinforcement Learning (HRL) has held longstanding promise to advance reinforcement learning. Yet, it has remained a considerable challenge to develop practical algorithms that exhibit some of these promises. To improve our fundamental understanding of HRL, we investigate hierarchical credit assignment from the perspective of conventional multistep reinforcement learning. We show how e.g., a 1-step `hierarchical backup' can be seen as a conventional multistep backup with $n$ skip connections over time connecting each subsequent state to the first independent of actions inbetween. Furthermore, we find that generalizing hierarchy to multistep return estimation methods requires us to consider how to partition the environment trace, in order to construct backup paths. We leverage these insight to develop a new hierarchical algorithm Hier$Q_k(\lambda)$, for which we demonstrate that hierarchical credit assignment alone can already boost agent performance (i.e., when eliminating generalization or exploration). Altogether, our work yields fundamental insight into the nature of hierarchical backups and distinguishes this as an additional basis for reinforcement learning research. 
\end{abstract}

\section{Introduction}  \label{sec:intro}
Hierarchical Reinforcement Learning (HRL) is often regarded as an open frontier in RL for developing more sample-efficient control algorithms \citep{sutton_between_1999, dietterich_maxq_1998, bakker_hierarchical_2004, sutton_reinforcement_2018, levy_learning_2018, kulkarni_hierarchical_2016}. Hierarchy provides innate structure for solving complex problems by decomposing tasks into smaller, simpler and recurring, subtasks. In turn, this allows the decision making algorithm to reason or plan over temporally distant events instead of only the (arbitrarily granular) environment actions.

A vast body of literature has documented and demonstrated potential benefits of hierarchy. There have also been a few impressive empirical results (e.g., \citep{jaderberg_humanlevel_2019}). Despite all this, it has remained difficult to design and work with hierarchical agents due to the additional challenges that these methods introduce. In essence, in HRL we often attempt to design algorithms that try to unite multiple, possibly non-stationary and unstable, policies into a more efficient whole. Research is occasionally hindered by issues relating to training instability, goal misspecification, or even a collapse of the hierarchy to a flat agent \citep{vezhnevets_feudal_2017, sutton_between_1999, nachum_dataefficient_2018}. While numerous methods effectively patch such encountered issues \emph{ad-hoc}, a better fundamental understanding of the learning dynamics will be crucial in advancing HRL research. 

The concept of learning in RL relates to the \emph{policy evaluation} problem: the estimation of the expected future return for a particular policy. Numerous successful return estimation algorithms have been proposed for flat agents, such as TD$(\lambda$) or Retrace \citep{sutton_reinforcement_2018, munos_safe_2016}. However, it is not directly obvious how these methods translate to the hierarchical setting. Hierarchical policies operate on varying levels of time granularity, and essentially can `jump' over flat actions when backing up rewards. 

We take as a practical example the hierarchical $Q$-learning algorithm by \citet{levy_learning_2018} and show how flat reward estimation methods can be adapted to hierarchy. We do this for the Tree-Backup operator \citep{sutton_reinforcement_2018} and an adaptation of Watkin's $Q(\lambda)$ \citep{watkins_qlearning_1992}, we generalize this into a new algorithm called Hier$Q_k(\lambda)$ where $k$ refers to the number of hierarchy levels. Finally, we analyze these methods in environment domains where we isolate the benefit of hierarchy to \emph{just} the credit assignment, in order to compare the hierarchical agents to similarly formulated flat approaches.

In short, we make the following contributions: 1) we systematically study hierarchical credit assignment patterns, and compare them to flat back-ups, 2) we propose a new algorithm, Hierarchical $Q(\lambda)$, which integrates these insights with Tree-Backup, and 3) we empirically compare the performance of this algorithm on a range of tasks, which shows hierarchy provides a fundamental performance benefit over flat agents through reward assignment alone (i.e., when removing other benefits of hierarchy provided by generalization, exploration, or transfer).

% notes:
% so that an agent may learn to swiftly return to previously experienced states. This can be useful for e.g., exploration, goal-directed behaviour can guide exploration towards states that are e.g., distant from the starting state --- states that would otherwise have low probability of being seen with a conventional exploratory policy (e.g., $\epsilon$-greedy).

\section{Related Work}  \label{sec:related}
The pursuit for Hierarchical structure in Reinforcement Learning agents has long been motivated by both philosophical considerations and promising results \citep{wen_efficiency_2020, dietterich_maxq_1998, kulkarni_hierarchical_2016, bakker_hierarchical_2004, pertsch_longhorizon_2020, nachum_dataefficient_2018} and is perhaps most well known under the Options framework of \citet{sutton_between_1999}. Though numerous RL algorithms have been designed that leverage benefits of hierarchy, such studies often put more emphasis in getting their method to work rather than to gain a better understanding of the underlying dynamics. 

Our work takes as a running example a recent hierarchical $Q$-learning algorithm by \citet{levy_learning_2018}, which illustrated strong performance in both discrete and continuous environments coupled with neural networks for function approximation. This method is similar to earlier work by \citet{bakker_hierarchical_2004}, the difference is the usage of neural networks and assuming that each observable state can be a goal state. Though their method can work well, it remained unclear what kind of benefits multiple levels of hierarchy could bring forth (as also mentioned by some of the reviewers in the OpenReview ICLR-2019 submission of \citet{levy_learning_2018}). This unclarity can be directly answered through our interpretation of hierarchical credit assignment.

A formulation for hierarchical agents that is similar to the multistep methods we develop in this paper is by \citet{jain_eligibility_2018}, who extended the option framework to general return estimation operators (such as, Retrace \citep{munos_safe_2016}, or Tree-Backup). They do this using an intra-option framework for updating their policies. However, their method was limited to the fact that individual levels could not train independently of one another, and it remained unclear whether their hierarchical structure actually provided any substantial benefits to conventional agents.

\section{Background}    \label{sec:background}
% Problem Statement
We consider solving Markov Decision Processes (MDP) which are 5-tuples $\setM=\langle \setS, \setA, p, R, \gamma \rangle$ where $\setS$ and $\setA$ are a set of states and actions; $p: \setS \times \setA \rightarrow \mathcal{P}(\setS)$ is a stationary transition distribution; $R$ maps transitions to rewards $R: \setS \times \setA \times \setS \rightarrow \doubleR$ which we abbreviate as $R_t = R(S_t, A_t, S_{t+1})$; and $\gamma \in [0, 1]$ is a discount factor \citep{sutton_reinforcement_2018}. Let $\pi: \setS \rightarrow \mathcal{P}(\setA)$ denote a control policy from which we can generate experiences\footnote{Notice that we make use of \emph{time-slicing} for sequences according to the Python convention. For example $\tau_t \equiv \tau_{t:T:k}$ means: the sequence $\tau$ starting at $t$ up until $T$ with jumps of $k$.} in $\setM$: $\tau_{t:T:k} = \{S_{t+ki}, A_{t+ki}, R_{t+ki+1}\}^{T-t}_{i=0}, k=1$. Then we seek the optimal policy $\pi^*$ that maximizes the value function, 
\begin{align}
    q_\pi(s, a)&=\doubleE_{\pi, p}[G_{t:T} | S_t=s, A_t=a] \label{eq:value} \\
    &= \doubleE_{\pi, p}[\sum^{T - t}_{i=0} \gamma^{i} R_{t + i+1} | S_t = s, A_t = a]  \nonumber,  
\end{align}
which, in practice, is estimated from samples $Q^\pi \approx q_\pi$.

% Q-learning
A famous algorithm that learns $q_\pi$ from data is $Q$-learning \citep{watkins_qlearning_1992}. At any transition $S_t, A_t, R_{t+1}, S_{t+1}$ it computes a biased estimation error using the recursive Bellman property of $Q^\pi$,
\begin{align}
    \delta_t = R_{t+1} + \gamma \max_a Q(S_{t+1}, a) - Q(S_t, A_t).  \label{eq:qlearning}
\end{align}
This error term is then used to adjust the current estimate to $q_\pi$ using the online update rule $Q^\pi = Q^\pi + \alpha \delta$, where $\alpha \in (0, 1]$ is a step-size parameter. Under some technical assumptions, $Q$-learning will converge to $\pi_*$ in the limit of infinite data \citep{watkins_qlearning_1992}.

% Multistep updates
\subsection{Multistep Backups}
Though $Q$-learning may converge to $\pi^*$ eventually, this method converges quite slowly as the estimation error $\delta$ is considered for $1$-step transitions only. More clever algorithms leverage experience gathered over multiple timesteps, e.g., Tree-Backup (TB$(n)$) generalizes $Q$-learning by updating towards
\begin{equation}
    \delta_{t:t+n} = \sum_{k=0}^{n - 1}\delta_{t+k} \prod_{i=1}^k \gamma \pi (A_{t+i} | S_{t+i}),  \label{eq:TB}
\end{equation}
which essentially computes $n$ single-transition errors $\delta_{t+k}$ and sums them according to their policy probabilities \citep{sutton_reinforcement_2018}. If we were to utilize a greedy target policy $\pi^*$, then Tree-Backup simply sums the rewards $n$ steps along the trace for as long as the actions are always greedy w.r.t. $Q^\pi$. 

For online learning, it should be obvious that TB$(n)$ also incurs a time-delay for updates as we need to wait for subsequent transitions. This consequence is inherent to this type of \emph{forward view} method: at $S_t$ we are looking $n$-steps ahead in time to construct $\delta_{t:t+n}$. It is also possible to express multistep updates using a \emph{backward view} where we cast only the \emph{current} estimation error $\delta_t$ towards our $Q^\pi$ estimates for previous experiences. Consider an eligibility trace $z(s, a) \in [0, 1], \forall (s, a)$ (by default set to zero), where we keep track of past transitions by updating the recency values,
\begin{equation}
    z(s, a) = \begin{cases} 1, \: &(s, a) = (S_t, A_t) \\ \gamma \lambda \pi(A_t | S_t) z(s, a), \: &(s, a) \ne (S_t, A_t) \end{cases}  \label{eq:eligibility}
\end{equation}
where $\lambda \in [0, 1]$ is a decay parameter that governs a bias-variance trade-of. This particular update rule for $z(s, a)$ is known as the \emph{replacing trace}. We can then update $Q^\pi$ with,
\begin{equation}
    Q^\pi(s, a) = Q^\pi (s, a) + \alpha \delta_t z(s, a) , \quad \forall (s, a)  \label{eq:qlambda}
\end{equation}
which yields the Tree-Backup$(\lambda)$ algorithm for general policy learning or Watkin's $Q(\lambda)$ for learning $\pi^*$ \citep{munos_safe_2016, sutton_reinforcement_2018}.

\section{Hierarchical Q-learning}  \label{sec:hierq}
% Problem Decomposition into Goals
Multistep reward algorithms considerably improve over $1$-step methods, yet, policy learning can still become arbitrarily slow. For example, when no rewards are observed or when Tree-Backup prematurely truncates backups when the behaviour policy diverges from the target policy \citep{munos_safe_2016}. Hierarchical $Q$-learning must handle \emph{credit assignment} differently due to its recursive structure that decomposes the full MDP task into distinct subtasks --- it must estimate returns for each task-specialized policy and handle multiple time resolutions. 

Each specialized policy is updated to maximize a pseudo-reward for reaching its goal state $s \in \setS$. Thus, instead of searching for one policy that maximizes $q_\pi$, we require a set of policies $\Pi = \{\pi_{\langle s_1 \rangle}, \pi_{\langle s_2 \rangle}, \dots, \pi_{\langle s_{|\setS|} \rangle}\}$ that each maximize their respective value function. We consider binary pseudo-rewards of the form $\textbf{r}_{t} \equiv \textbf{1}_{S_{t}}$,  which is also known as the Successor Representation pseudo-reward \citep{dayan_improving_1993}. Alternatively, we could write this as a vector of zeros with a one at the index corresponding to state $S_{t}$, i.e., the Kronecker delta  $(\textbf{r}_t)_i = 1_{S_i = S_t}$. As a result, we are guaranteed to observe a reward (intrinsically) at every state transition. So, when we do observe an environment (extrinsic) reward, we can leverage the knowledge contained within $\Pi$ to follow this new gradient of interest. 

\begin{figure}  % TODO: Left align Pi_2
    \centering
    \includegraphics[width=\linewidth]{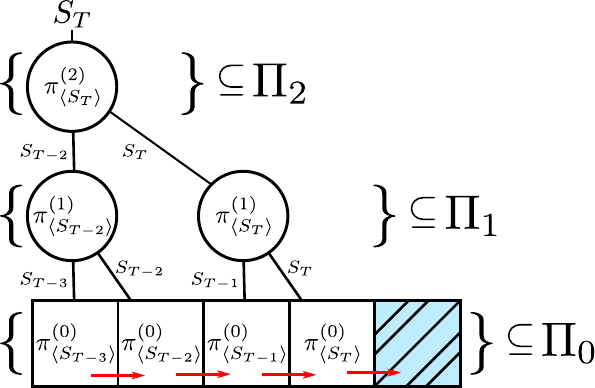}
    \caption{Example structure of the hierarchical task decomposition with $k=3$ levels of hierarchy. Given the final goal $S_T$, the top level $i=2$ sequentially samples the two sub-policies at $i=1$ that effectively stipulate a goal-trajectory towards $S_T$. These sub-policies in turn perform a similar procedure, but conditioned on a different sub-goal $\{S_{T-2}, S_T\}$ and on a more granular time-scale. Finally, the flat $i=0$ policies take environment actions to achieve their instructed goal (red arrows). }
    \label{fig:hierarchy_tree}
\end{figure}

% UAI FORMAT
\begin{algorithm}[!ht] 
    \SetKwProg{Fn}{Function}{\string:}{}
    \DontPrintSemicolon
    \SetKwInOut{Input}{Input}
    \SetKwFunction{FRecurse}{Recurse}%
    \caption{Hierarchical policy training procedure.} \label{alg:hierQrecursion}
    \Input{Hierarchy size $k$ and horizon $H_i$ for each level $i$}
    \BlankLine
    Initialize $\pi_i$ for each level $i = 0, \dots, k - 1$\;
    Initialize $S$ and $S_{\text{goal}}$ \;
    \While{$S \ne S_{\text{goal}}$}{
        $S \gets$ \FRecurse($k-1$, $S$, $S_{\text{goal}}$) \;
    }
    \BlankLine
    \Fn{\FRecurse($i$, $S$, $S_{\text{goal}}^{(i)}$)}{
        Set counter $n \gets 0$ \;
        \While{$n < H_i$ $\normalfont{\textbf{and}}$ $S \ne S_{goal}^{(j)},\forall j, j \ge i $}{
            Sample action $A^{(i)} \sim \pi_i(a \: | \: S, S_{\text{goal}}^{(i)})$ \;
            \eIf{$i > 0$}{
                $S' \gets$ \FRecurse($i - 1$, $S$, $A^{(i)}$) \;
            }{
                Apply $A^{(0)}$, observe transition $S', R$ \;
                Update $\Pi_i, i = 0, \dots, k-1$ \: \{e.g., Alg. \ref{alg:HierQLambda}\} \;
            }
            $n \gets n + 1$, $S \gets S'$ \;
        }
        \textbf{return} $S$ \;
    }
\end{algorithm}

From the structure of $\Pi$ we can define policies over policies (akin to Options \citep{sutton_between_1999}) in order to sample goals to reach, the agent can then stipulate a trajectory of goals that maximize the environment reward (see Figure~\ref{fig:hierarchy_tree}). Hierarchical $Q$-learning does this in recursive fashion, as depicted in Figure~\ref{fig:hierarchy_tree}  and Algorithm~\ref{alg:hierQrecursion}. The method creates a tree of $k$ goal-sampling policies $\Pi_i, i = 0, \dots, k-1$ that each instruct their lower level\footnote{For notation, we denote a hierarchy level $i$ with subscripts. For overloaded indices (e.g., time), we use superscripts $(i)$.} (direct child node) to achieve some state in finite time (e.g., in $H_i$ steps). In other words, for any $\pi_i \in \Pi_i$, its action space is $\setA_i \subseteq \Pi_{i-1}, i > 0$, with $\setA_0 \equiv \setA$, and its episode horizon is $H_i$. The hierarchy prunes branches when nodes exceed their budget $H_{i}$ or when a goal state is achieved (at any level). As a result, any level $i>0$ is \emph{semi-Markov} with a maximum atomic horizon (the time-span of executed actions) of $H^a_i = \prod_{j=0}^{i - 1} H_j$ \citep{sutton_between_1999}. For $k=1$ we collapse to a flat agent, making this formulation a recursive generalization to flat RL.

\subsection{Hierarchical Policy Evaluation}
A desirable property of Algorithm \ref{alg:hierQrecursion} is that it allows us to update \emph{all} policies (that is for each hierarchy level) after every environment step (line~13). In contrast to conventional update mechanisms (such as TB$(\lambda)$), hierarchical actions do not have to be completed when performing updates. \citet{levy_learning_2018}, showed that we can relabel hierarchical actions (in hindsight) as the states that were reached rather than those that were instructed. This philosophy leads to a dense and counterfactual learning mechanism in the sense that: if the current state had been an instructed goal, then this \emph{would have been} a valid \emph{hierarchical} action at previously encountered states (and optimal if the current state was a goal state).

\begin{figure}[t]
    \centering
    \includegraphics[width=\linewidth]{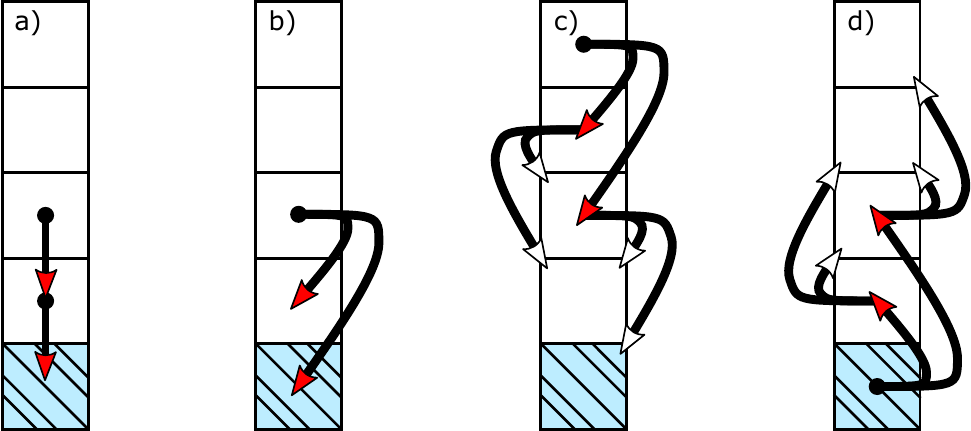}
    \caption{Comparison of a flat $n=2$ backup at $i=0$ (a), a hierarchical $n=1$ and $H^a_1=2$ backup at level $i = 1$ (b) and the set of all \emph{possible} backups for a hierarchical $n=H^a_1=2$ backup (c) along with its backward view (d) both at level $i=1$. The backups are all conditioned on, and all utilize a shared trace towards, the light-blue, dashed, goal-tile $S_T$. White arrows indicate that the backup paths are inferred through partitioning of the trace (multistep) whereas the red arrows indicate states that are reachable for the hierarchy (1-step) starting from the reference state (black dots). Heuristically, we could sparsify the \emph{white} arrows by only considering those with maximum time-span: $H^a_i = 2$. }
    \label{fig:backup}
\end{figure}

% Interpretation of the backup
The general idea is illustrated in Figure \ref{fig:backup} in a \emph{forward view} for (a, b, c) and a \emph{backward view} in (d). Given some environment trace, the conventional way of looking at a multistep ($n$-step; Equation \ref{eq:TB}) backup is to consider $n$ environment actions being applied sequentially (Figure \ref{fig:backup}a). In contrast, the hierarchical action relabelling implies that, counterfactually, a state $S_t$ \emph{could} have been a goal-action at each of the preceding $H^a_i$ states--- i.e., the states that lie within the hierarchical policy's atomic horizon. As an example, for $H^a_i=2$, a $1$-step update implies that the set $\{S_{t-1}, S_{t-2}\}$ contains all valid preceding states for the action $S_{t}$.
\begin{quote}
    \centering
    \emph{Hierarchical agents can update on paired state events $(S_t, S_{t+j})$, ignoring events inbetween.}
\end{quote}  % Density of backups and sparsification.
When we consider hierarchical $n$-step backups through this lens, we can see that the number of backup paths grows quite swiftly (as shown for $n=2$ in Figure \ref{fig:backup}c; now, there are $4$ backup paths of length $n$). We can actually show that the number of all possible backup paths grows super-exponentially with respect to $k$, $n$, $H_i$ and $t$ (for more details, see Appendix~\ref{ap:partition}). Hence, for online learning, considering all possible backup paths will quickly become infeasible, however, most of these paths will show overlap. As illustrated by the backward view in Figure \ref{fig:backup}d, looking back from the goal-tile, it doesn't make much sense to perform a $n=2$ backup from $S_{T-1}$ to $S_{T-2}$ when $S_{T-2}$ lies within the $H^a_i$ horizon, we can jump directly from $S_T$ to $S_{T-2}$. 

For this reason, we suggest to sparsify the multistep backups by only considering those with maximum valid length. To view the effect of this sparsification in Figure~\ref{fig:backup}c,d, imagine pruning all white arrows that do not make time-jumps of length $H^a_i$ (maximum valid time-span). This is somewhat heuristic of course, but it makes the return computation tractable and incentivizes the hierarchical policies to take actions that make maximal use of their action-budget. Naturally, if we'd consider backup paths of $H^a_i=1$, we would end up with a flat backup and effective policy.

\subsection{Algorithm Formulation}
Denote $k$ estimates for $q_\pi$ at each level in the hierarchy as $Q_i \subset \mathbb{R}^{|\setS|\times |\setA_i| \times |\setS|}, i=0,\dots, k-1$. Then, observe that for a hierarchical policy $\pi_i \in \Pi_i, i > 0$ at state $S_t$ we can look forward on the trace to find the set of reachable actions $\textbf{A}^{(i)}_t=\{S_{t+j}\}_{j=1}^{H^a_i}$. If we extend the $1$-step update from Equation \ref{eq:qlearning} to consider the hierarchical time skips and all specialized goal policies contained in $\Pi_i$, we get:
\begin{align}
    \bm{\delta}_{t}^j &= \textbf{r}_{t+j} + \bm{\gamma}_{t+j} \mathbb{E}_\pi Q_i(S_{t+j}, \cdot, \textbf{g}) - Q_i(S_t, A^{(i)}_{t+j - 1}, \textbf{g}),  \nonumber \\
    &= \vecG_t^{j} -  Q_i(S_t, A^{(i)}_{t+j - 1}, \textbf{g}) \label{eq:hierQ}
\end{align}
where $A^{(i)}_{t+j - 1} \equiv S_{t+j}, i > 0$ and $\mathbb{E}_\pi Q_i$ is short-hand for the expectation of $Q_i$ under the target policy $\pi$. Note that $\bm{\delta}_t^j \in \doubleR^{|\setS|}$ is now a vector of errors for each policy $\pi_i \in \Pi_i$, this is made explicit by the goal-vector $\textbf{g}$. Accordingly, the vector $\textbf{r}_{t+j} = (1 - \bm{\gamma}_{t+j})/\gamma = \bm{1}_{S_{t+j}}$ is the state indicator as discussed before and bold $\bm{\gamma}$ is a termination function that ends (restarts) episodes when a $S_{t}$-specialized policy achieves $S_t$. If we then map the set of state-action pairs $\{S_t\}\times\textbf{A}_t^{(i)}$ to the set of corresponding update targets $\Delta_t^{(i)}=\{\bm{\delta}_t^1, \dots, \bm{\delta}_t^{H^a_i}\}$ using a greedy target policy $\pi^*$, we get the update target for Hier$Q_k$ \citep{levy_learning_2018}.

\subsubsection*{Hier$\text{TB}_k(n)$}  \label{sec:tb}
We can extend the single step update targets in $\Delta_t^{(i)}$ to multistep ones through the Tree-Backup operator. To reduce clutter, denote $h=H^a_i$, then we can write,
\begin{align}
    \bm{\delta}_{t:t+nh:h}^j &= \sum_{k=0}^{n-1}\bm{\delta}_{t+kh}^j \prod_{l=1}^k \bm{\gamma}_{t+j+lh} \pi (A^{(i)}_{t+j+lh} | S_{t+j+lh}, \textbf{g}) \nonumber \\
    &=  \sum_{k=0}^{n-1}\bm{\delta}_{t+kh}^j \prod_{l=1}^k \bm{\gamma}_{t+j+lh} \bm{\pi}_{t+j+lh}, \label{eq:HTB} 
\end{align}
with $A^{(i)}_{t+j+lh} \equiv S_{t+j+(l+1)h}, i > 0$, as the general HierTB$_k(n)$ error for each $\Pi_i, i=0, \dots, k-1$ (with corrected $h$) and for any well-defined target policy $\pi$. Our sparsification can be observed from the time-jumps towards the farthest allowed state on the trace from each reference state, i.e., from $S_t$ towards $S_{t+H^a_i}$ as indicated by the subscripts (see also Figure \ref{fig:backup}c,d for a visual reference; all white arrows with span $<H^a_i$ get pruned). Our implementation of Tree-Backup performs hierarchical updates with a similar time complexity as conventional Tree-Backup. This is a drastic improvement over any naive implementation, which would be of exponential time-complexity (see Appendix~\ref{ap:sec:tb_backup}). 

\subsubsection*{Hier$Q_k(\lambda)$}  \label{sec:hql}

% UAI FORMAT
\begin{algorithm}[tb] 
    \DontPrintSemicolon
	\caption{Hierarchical $Q(\lambda)$/ Hier$Q_k(\lambda)$} \label{alg:HierQLambda}
    \nonl \textbf{Input :} Environment trace $\tau_{t+1} = \{S_j, A_j\}_{j=0}^{t+1}$, the hierarchy level $i$, eligibilities $Z^{(i)}$, a target policy $\pi$, and $Q$-table $Q_i(s, a, g)$ \; \nonl \textbf{Parameters :} Discount factor $\gamma \in [0, 1)$, step-size $\alpha \in (0, 1]$, decay $\lambda \in [0, 1]$, and policy reach $H^a$
    \BlankLine
    Let $h \gets t \mod H^a$ and $t_{\min} \gets \min(H^a - 1, t)$ \;
    Infer level action $A \gets A^{(0)}$ or $S_{t+1}$ \;
    $\textbf{G} \gets \textbf{r}_{t+1} + \bm{\gamma}_{t+1}\mathbb{E}_\pi Q_i (S_{t+1}, \cdot, \textbf{g})$ \;
    $\bm{\delta} \hspace{0.24em} \gets \textbf{G} - Q_i(S_{t - t_{\min}}, A, \textbf{g})$ \;
    $Z^{(i)}_{h}(s, a, \textbf{g}) \gets \lambda \bm{\gamma}_{t - t_{\min}} \bm{\pi}_{t - t_{\min}} Z^{(i)}_h (s, a, \textbf{g}), \: \forall (s, a)$ \;
    $Q_i (s, a, \textbf{g}) \hspace{.55em} \gets Q_i(s, a, \textbf{g}) + \alpha \bm{\delta} Z^{(i)}_h(s, a, \textbf{g}), \hspace{\fill} \forall (s, a)$  \;
    \For{$j = 0, \dots, \min(H^a - 1, t_{\min})$}{
        $Z^{(i)}_h (S_{t_{\min} - j}, A, \textbf{g}) \gets 1$ \;
        $Q_i(S_{t_{\min}-j}, A, \textbf{g}) \hspace{.6em}\gets (1-\alpha)Q_i(S_{t_{\min}-j}, A, \textbf{g}) + \alpha \textbf{G}$ \;
    }
\end{algorithm}

% ICLR FORMAT
% \begin{algorithm}[H] 
%     \DontPrintSemicolon
% 	\caption{Hierarchical $Q(\lambda)$/ Hier$Q_k(\lambda)$} \label{alg:HierQLambda}
%     \nonl \textbf{Input :} Environment trace $\tau_{t+1} = \{S_j, A_j\}_{j=0}^{t+1}$, the hierarchy level $i$, eligibilities $Z^{(i)}$, a target policy $\pi$, and $Q$-table $Q_i(s, a, g)$ \; \nonl \textbf{Parameters :} Discount factor $\gamma \in [0, 1)$, step-size $\alpha \in (0, 1]$, decay $\lambda \in [0, 1]$, and policy reach $H^a$
%     \BlankLine
%     Let $h \gets t \mod H^a$, $t_{\min} \gets \min(H^a - 1, t)$ \;
%     $A \gets A^{(0)}$ or $S_{t+1}$ if $i > 0$ \;
%     $\textbf{G} \gets \textbf{r}_{t+1} + \bm{\gamma}_{t+1}\mathbb{E}_\pi Q_i (S_{t+1}, \cdot, \textbf{g})$ \;
%     $\bm{\delta} \hspace{0.24em} \gets \textbf{G} - Q_i(S_{t - t_{\min}}, A, \textbf{g})$ \;
%     $Z^{(i)}_{h}(s, a, \textbf{g}) \gets \lambda \bm{\gamma}_{t - t_{\min}} \bm{\pi}_{t - t_{\min}} Z^{(i)}_h (s, a, \textbf{g})$ \;
%     $Q_i (s, a, \textbf{g}) \hspace{.55em} \gets Q_i(s, a, \textbf{g}) + \alpha \bm{\delta} Z^{(i)}_h(s, a, \textbf{g}), \hspace{\fill}$  \;
%     \For{$j = 0, \dots, \min(H^a - 1, t_{\min})$}{
%         $Z^{(i)}_h (S_{t_{\min} - j}, A, \textbf{g}) \gets 1$ \;
%         $Q_i(S_{t_{\min}-j}, A, \textbf{g}) \hspace{.6em}\gets (1-\alpha)Q_i(S_{t_{\min}-j}, A, \textbf{g}) + \alpha \textbf{G}$ \;
%     }
% \end{algorithm} 

Let us first define the eligibility trace as the matrix $Z^{(i)} \subseteq [0, 1]^{|\setS| \times | \setA_i| \times | \setS|}$ for each level $i = 0, \dots, k-1$. So, each column in $Z^{(i)}$ tracks an individual eligibility trace (c.f., Equation~\ref{eq:eligibility}) for every separate policy in $\Pi_i$. Denote the set of valid preceding states $\textbf{S}^{(i)}_t = \{S_{t-j}\}_{j=0}^{H^a_i}$ with valid action $A^{(i)}_t$, we can then write the update for $Z^{(i)}$, for each element $(s, a) \in \Omega \equiv \textbf{S}^{(i)}_t \times \{A^{(i)}_t\}$, according to,
\begin{equation}
    Z^{(i)}(s, a, \textbf{g}) = \begin{cases} 1, \: &(s, a) \in  \Omega \\ \lambda \bm{\gamma}_k\bm{\pi}_k Z^{(i)}(s, a, \textbf{g}), \: &\text{Otherwise} \end{cases} \label{eq:hiertrace}
\end{equation}
where $\bm{\pi}_k$ contains the transition (state-action) probabilities for each goal, $\bm{\gamma}_k$ terminates achieved state-goals, and $k \in \{t, t-1, \dots, t-H^a_i + 1\}$. Like before (the sparsification from hierarchical Tree-Backup) we can assume that the policy always takes actions with maximal temporal span $\bm{\pi}_k \implies \pi(A_t^{(i)} | S_{t+1 - H^a_i}, \textbf{g})$. 

It turns out, to get an exact generalization of our version of Hierarchical TB$(n)$ to hierarchical TB$(\lambda)$, we need to keep track of $H^a_i$ disjoint eligibility matrices $Z^{(i)}_h$ for $h=1, \dots, H^a_i$ in order to correctly track the sparsified back-up paths for each hierarchy level. We can then circulate through each eligibility matrix by only utilizing the eligibility $Z^{(i)}_h, h = {t \mod H^a_i}$ at timestep $t$. This idea is illustrated in Algorithm~\ref{alg:HierQLambda}, which in the case of a greedy target policy $\pi$ yields a hierarchical generalization of Watkin's $Q(\lambda)$, which we dub Hier$Q_k(\lambda)$. In the pseudocode, we keep the 1-step updates separate from the eligibility trace update initially (line 9; Algorithm~\ref{alg:HierQLambda}), seeing as each element $(s, a) \in \Omega$ induces a different error $\bm{\delta}$ even though they share the same returns $\textbf{G}$. Of course, this is only relevant for the 1-step errors, after which these trailing pairs are simply added to the eligibility trace (line 8; Algorithm~\ref{alg:HierQLambda}).

\section{Empirical Evaluation}  \label{sec:experiments}
From our formulation of the hierarchical backup, along with their algorithmic implementations, we can see how hierarchical credit assignment differs from conventional (flat) credit assignment in a number of ways. Most notably, the hierarchy induces skip-connections over the environment trace when computing returns. As a result, we don't need to account for correction terms at each environment transition (without decaying, discounting, or truncation), which allows us to propagate rewards much further back in time. 

\begin{figure*}[!t]
    \centering
    \includegraphics[width=.9\linewidth]{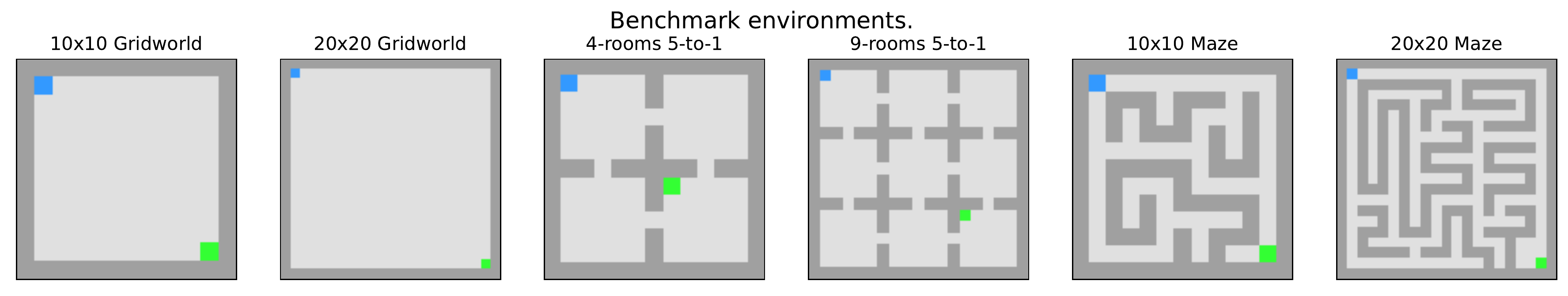}
    \caption{Overview of the tested environments. The agent starts out at the blue tile and is rewarded at the green tile.}
    \label{fig:ablations_envs}
\end{figure*}

It is a well known phenomenon that conventional Tree-Backup (or Watkin's $Q(\lambda)$) truncates traces too often due to possible divergence between behaviour policies and the target policy \citep{munos_safe_2016, peng_incremental_1996, kozuno_revisiting_2021}. Naturally, this trace truncation is for good reason from a conservative perspective: we would otherwise be estimating a different policy. A similar reasoning applies to the $\lambda$ parameter for the eligibility trace and the parameter $n$ for the multistep return, which both interpolate between having a more biased estimate of the return or one with potentially high variance. When rewards are propagated far backwards in time, the resulting update can become arbitrarily noisy and unstable.

All in all, this may raise the question whether a hierarchical backup only yields a benefit (or detriment) to the agent through deeper reward propagation (see also Appendix~\ref{ap:policy_example} for a direct comparison of a flat and hierarchical agent trained on the same trace). Hypothetically, we could achieve a similar effect of deeper reward propagation by just increasing $n$ or $\gamma, \lambda$ for the flat agent (not exactly of course, due to the aforementioned truncation issue). Thus, we ran experiments to analyze the difference in behaviour and performance of these agents over various backup parameters.

\subsection*{Experimental Setup}
We evaluated the Hierarchical Tree-Backup and $Q(\lambda)$ implementations from Algorithm~\ref{alg:HierQTB} and Algorithm~\ref{alg:HierQLambda} over two separate parameter grids on each of the discrete gridworld environments portrayed in Figure~\ref{fig:ablations_envs} (See also Appendix~\ref{ap:experiment} for additional experiment details). Each environment was chosen to provide either a distinct structure to the state-action spaces or to their scale. We opted for a tabular setting to eliminate most other benefits posed by hierarchy \citep{wen_efficiency_2020}, e.g., to eliminate possible generalization between goals or goal-directed exploration. The agents were initialized (with $Q_i = \textbf{0}, \forall i$) at the blue-tile and could traverse the environment deterministically with the actions $\setA^{(0)} = \{\uparrow, \downarrow, \leftarrow, \uparrow\}$. Upon reaching the green-tile, the agents received a sparse binary reward. 

The first, and largest, parameter grid evaluated our algorithms over various hierarchy levels $k \in \{1, 2, 3, 4\}$, backup steps $n \in \{1, 3, 5, 8\}$, decay values $\lambda \in \{0, 0.5, 0.8, 1\}$, and behaviour policies $\{\Pi_{k-1}, \pi_0\}$ (only during training). This study aimed to quantify the marginal performance benefit of evaluating and training with additional hierarchy for various reward backup depths. The second parameter grid was evaluated for $k \in \{1, 2, 3\}$, behaviour policies $\{\Pi_{k-1}, \pi_0\}$ (only during training), and decay $\lambda = 1$, with derived parameters $\gamma = (\gamma_0)^{1/H^a_{k-1}}$ and $n = n_0 / H^a_{k-1}$. These formulas were intended to adjust the discount and backup parameters such that every hierarchy level $k$ in this ablation study sent credits back equally far. For a complete overview of all parameters and their descriptions, see Table~\ref{fig:flat_ablations_mean} in the supplementary material. We utilized $\epsilon$-greedy exploration with $\epsilon=0.25$ for all flat policies $k=1$ ($i=0$) during training and $\epsilon=0.05$ during evaluation. All hierarchical polices utilized a fully greedy policy $\epsilon = 0$ (uniform tie-breaking) such that exploration was mostly handled by the flat level. 

This choice for hierarchical exploration is well motivated in our case seeing as uniform random exploration compounds on each level, $Pr(\text{Random-Policy}) = 1 - (1 - \epsilon)^k$. Considering the fact that hierarchical policies sample actions that carry over multiple time-steps, uniformly random exploration would result in erratic behaviour of the effective environment policy --- especially as the dimensionality of the action-space increases. Of course we could have opted for a different exploration policy (e.g., Boltzmann exploration), but this would have more strongly confounded our results due to more efficient hierarchical exploration.

All agent configurations were evaluated over 200 random seeds on each environment utilizing a simple alternating train-test loop for $50$ iterations (after which most agents had converged). Training episodes were terminated and reset when the agent exceeded a budget of $10^5$ steps, this was sufficiently large to let every agent configuration observe a reward in their first episode. We aggregated the test-performance, measured in the number of $\log(\text{steps})$ needed to reach the green-tile starting from the blue-tile, over all repetitions per time-step to produce loss-curves. We chose to log-transform the step-counts to stabilize the variance of the means. Finally, we averaged over the loss-curves to quantify the marginal performance, this could be interpreted as an unnormalized area under the \emph{mean} performance curve.

\begin{figure*}[t]
    \centering
    \includegraphics[width=.98\linewidth]{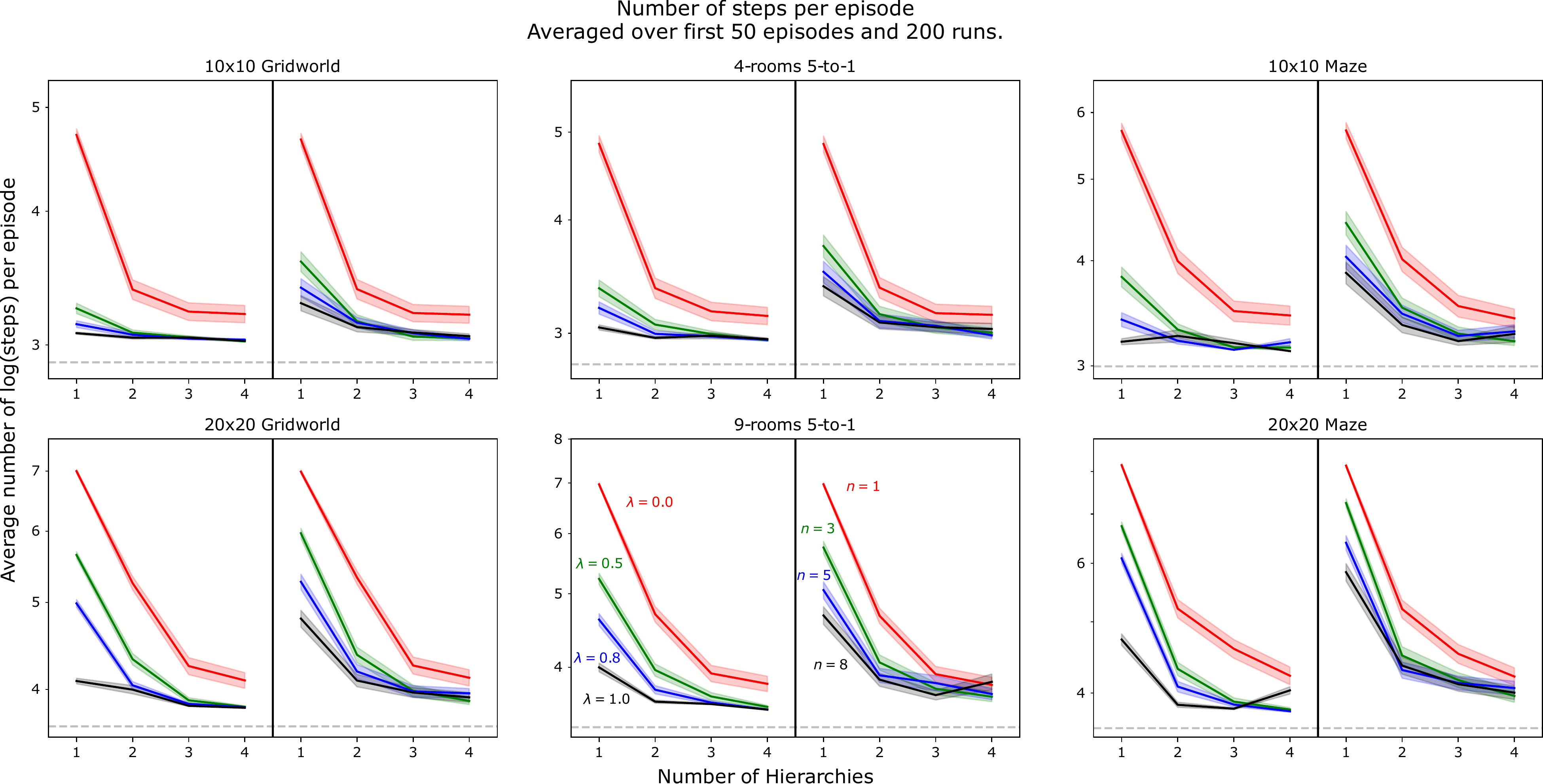}
    \caption{Marginal log-performance of each experiment configuration for each environment from Figure~\ref{fig:ablations_envs} (lower is better). All agents utilized a base-discount of $\gamma = 0.95$. The configurations are split left for the eligibility trace agents (Algorithm~\ref{alg:HierQLambda}) and right for Tree-Backup agents (Algorithm~\ref{alg:HierQTB}) as indicated by the colored annotations. Shaded regions indicate \textonehalf-standard errors of the mean and grey-dashed lines indicate optimal performance per environment. }
    \label{fig:ablations_mean}
\end{figure*}

\begin{table}[!th]
    \centering
    \renewcommand{\arraystretch}{1.1}
    \caption{Example distribution statistics for the trace length of the first training episode (the first observed reward). The mean $\mu$ can roughly be interpreted as the scale $\beta$ of an exponential distribution, $\text{Exp}(\beta^{-1})$. Additional hierarchy levels seem to prolong the first episode in the $20 \times 20$ Gridworld, however this pattern is not monotonic and seems to reverse at $k>2$ on the $10 \times 10$ Maze. } \label{tab:first_reward}
    \vspace{-1.5em}
    \bigskip

\centering
\begin{tabular}{p{1.3cm}|l|p{2cm}|p{2cm}}
    Backup $n$ or $\lambda$         &   Levels $k$          & $\mu \pm s^-$: \newline $20 \times 20$ \newline Gridworld  & $\mu \pm s^-$: \newline$10 \times 10$ Maze  \\
    \hline
    $n=3$           &   $k=1$       &   $3064 \pm 210$      &   $2493 \pm 171$      \\
                    &   $k=2$       &   $5176 \pm 394$      &   $5000 \pm 555$      \\
                    &   $k=3$       &   $8010 \pm 600$      &   $2573 \pm 247$      \\
                    &   $k=4$       &   $6563 \pm 628$      &   $1681 \pm 143$      \\
    \hline
    $\lambda = 0.8$ &   $k=1$       &   $3300 \pm 189$      &   $2525 \pm 187$      \\
                    &   $k=2$       &   $6528 \pm 436$      &   $6318 \pm 665$      \\
                    &   $k=3$       &   $8254 \pm 663$      &   $2914 \pm 267$      \\
                    &   $k=4$       &   $8406 \pm 788$      &   $1663 \pm 143$      \\
    
\end{tabular}
\end{table}

\subsection*{Experiment Results}
The mean results (with standard errors) for the first parameter grid, conditioned on the hierarchical training policy, are visualized fully in Figure~\ref{fig:ablations_mean}. Generally speaking, the pattern indicates that hierarchy near-monotonically improves upon its flat counterpart $k=1$ for any parameter setting (in expected log-score). Though, hierarchy generally did indeed improve upon the flat counterparts, it also exhibited much higher variance; in the first number of episodes we found that hierarchy could actually degrade performance. Regularly, the hierarchical agents would take a marginally longer time to finish their first few training episodes compared to the flat agents. This is also substantiated by the results in Table~\ref{tab:first_reward} (and Table~\ref{tab:appendix:first_reward} in the Supplementary Material). The table indicates that generally, additional hierarchy levels prolong the first episode. However, this pattern occasionally jumps back when the hierarchy level is taken to an extreme, which is especially visible on the $10 \times 10$ Maze environment.

It is also noteworthy that the flat $\lambda = 1$ agents consistently performed in a competitive manner to the hierarchical policies, especially on the small environments. The environments were deterministic after all, and these agents were capable of sending rewards back the farthest of the flat policies. As also noted in Table~\ref{tab:appendix:first_reward}, these agents generally spent the least amount of time wandering about in the first episode.

% Note: make clear that data-generation is different, and highlight that keeping this uniform will improve the agent.
To analyze the effect of reward backup depth on the quality of the resulting policy, it must be noted that the results in Figure~\ref{fig:ablations_mean} are confounded by the fact that the training data was generated by entirely different policies for the flat and hierarchical agents (as should be evident from Table~\ref{tab:appendix:first_reward}). For a more fair comparison to the effect of reward backup depth, Figure~\ref{fig:loss_curve} (see also Figures~\ref{fig:full_loss_curve_n},\ref{fig:full_loss_curve_lambda} in the Supplementary Material) visualizes the mean performance of the agents over time/ training episodes for two of the tested environments (chosen for visual clarity). Still, this result shows a similar pattern as observed in Figure~\ref{fig:ablations_mean}: even with appropriately balanced backup depths and training either with or without the hierarchical structure, the positive effect on performance due to hierarchy seems to persist. This result also shows that the test scores for the hierarchical agent trained with the \emph{flat} behaviour policy improved marginally slower in the large Gridworld and faster in the small Maze environments. We also see that the flat agent $k=1$ (red line) performed worse during evaluation than both hierarchical policies. 

\begin{figure}[t]
    \centering
    \includegraphics[width=\linewidth]{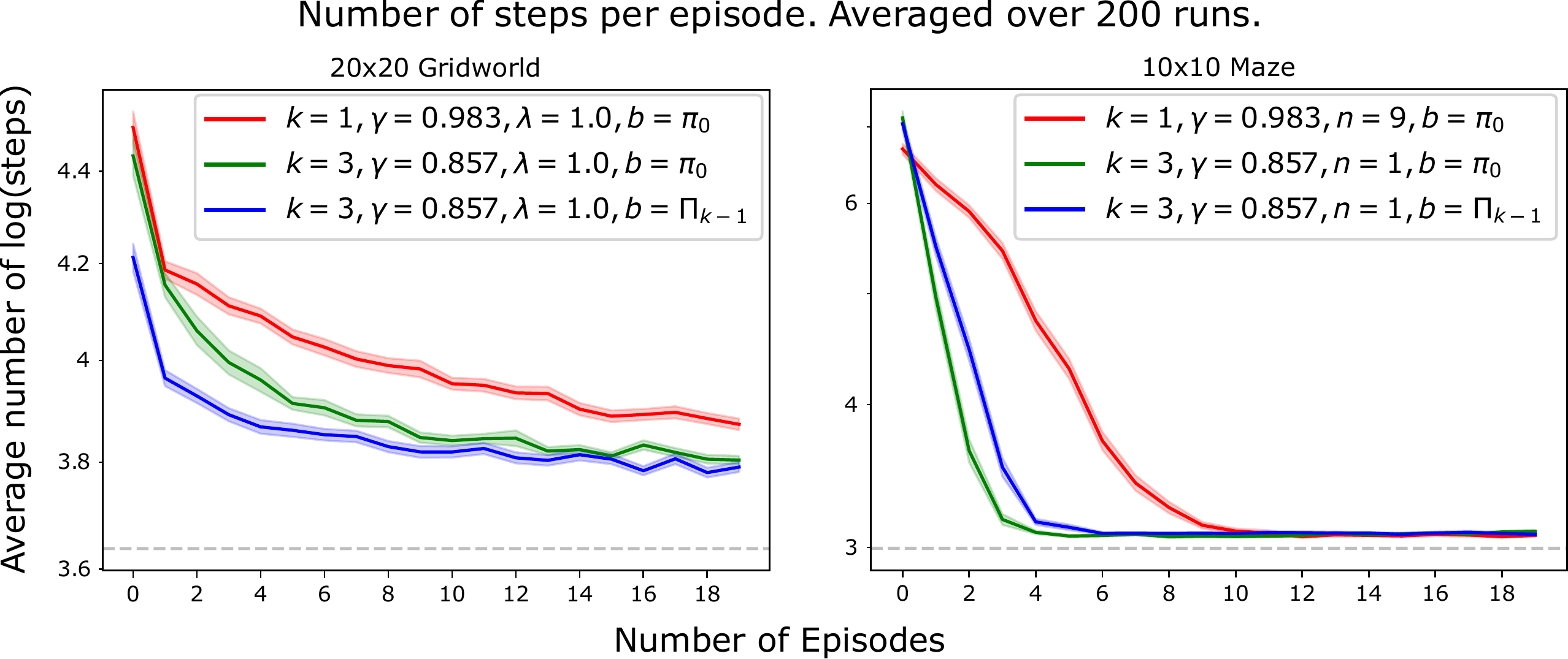}
    \caption{Average $\log$(score) of the evaluated agents at each training episode (lower is better) when credit assignment \emph{depth} is appropriately balanced for each hierarchy level $k$ and when training data is generated with either a flat policy $b=\pi_0$ or with the full hierarchy $b=\Pi_{k-1}$. Data for the hierarchical agent was generated with two different policies to account for the observation that hierarchy could prolong episodes (as noted in Table \ref{tab:first_reward}). Shaded regions indicate $1$-standard error of the mean.}
    \label{fig:loss_curve}
\end{figure}

\section{Discussion}    \label{sec:discussion}
Though our results may hint that hierarchy supersedes conventional, flat, agents, they often exhibited larger variance in the first few episodes. Also rewards were not always observed as swiftly by the hierarchical agents compared to the flat agents. These results are actually an artifact of our choice of policy parameterization, $\epsilon$-greedy for level $i=0$ and fully greedy for $i>0$. For the conventional agents, as long as no reward was observed the agent would do an uniform random walk. For the hierarchical agent, this was also the case for $i = k-1$ (the top-level). However, all levels below $k-1$ would swiftly learn to follow their instructed goals as a result of their state-specialization; they became experts in reaching the proposed goals by level $i=k-1$ even though this agent was still random. As a result, the agent became biased towards the states for which it could quickly accumulate knowledge. The top-level could then randomly sample a state, far from the goal-state, and the levels below would naively, skillfully, follow.

This issue could of course have been handled through alternative hierarchical exploration methods, however, for a fair comparison to the conventional agent we chose to simply generate training data with a flat policy (as shown in Figure~\ref{fig:flat_ablations_mean} in the Appendix). This again showed that hierarchy generally improves upon the flat agents, though the effect seemed to become less pronounced with additional hierarchy levels. In a way, our results should emphasize that $\epsilon$-based exploration becomes more and more unreliable as we decrease the temporal resolution. To effectively leverage the benefits posed by hierarchy, agents should follow a more principled way for exploration (c.f., \citep{kulkarni_hierarchical_2016, bellemare_unifying_2016, ecoffet_first_2021}). 
 
When we again juxtapose the definitions of flat and hierarchical multistep backups, the main distinctions that we can make are: how the rewards are sent back and how the trajectory is partitioned in order to perform updates. As mentioned before, Tree-Backup and its variants compute policy correction terms over state-transitions to estimate the returns for good reason. The return would otherwise be estimating a different policy, leading to e.g., a SARSA update \citep{sutton_reinforcement_2018}. Anecdotally, initial experiments showed that SARSA (or variants thereof) always performed worse than $Q$-learning in our experimental setting. Thus, it is interesting that hierarchy does not degrade performance in a similar way, but rather improves upon it. Possibly, we could view hierarchical structure to make $Q$-learning (or any other conservative return operator) amenable to non-conservative updates --- i.e., akin to SARSA or Peng's $Q(\lambda)$ \citep{sutton_reinforcement_2018, peng_incremental_1996, pertsch_longhorizon_2020}. 

Of course, we should not attribute the performance benefit of hierarchy merely to the update mechanism, but also to its innate structure. The recursive decomposition induces a form of `stitching effect' between policies that contain some knowledge about the task. However, it is not clear if this effect can be measurably distinguished from the hierarchical backups.

Finally, it must be noted that our description for Hier$Q_k(\lambda)$ does not yield a practical algorithm. Memory does not scale well in the tabular setting, especially when considering that we track $|\setS|$ policies at each level. The same reasoning goes for the eligibility traces, which in the worst-case scales proportional to $O(H^a_{k-1}|\setS|^2|\setA|)$, seeing as $Z \in \doubleR^{|\setS| \times |\setA^{(i)}| \times |\setS|}$ for which we keep $H^a_{k-1}$ copies. Our implementation adopts some heuristic optimizations in order to make our experiments tractable (see Appendix~\ref{ap:heuristics}), further optimizations such as a LIFO buffer or approximations to the eligibilities are left as future work. Our version of Hierarchical Tree-Backup should be readily applicable to hierarchical $Q$-learning methods even if e.g., (non-linear) function approximation is used \citep{levy_learning_2018}.

\subsection*{Conclusion}
This paper takes a fundamental perspective on how credit assignment is performed in Hierarchical Reinforcement Learning through a direct comparison to multistep return estimation methods \citep{sutton_reinforcement_2018}. We take as a running example the Hierarchical $Q$-learning method by \citet{levy_learning_2018} to illustrate how such an algorithm may construct estimations of its returns. Our perspective showed that the number of possible ways to construct update targets for hierarchical policies grows rapidly over time but can be sparsified by focusing only on a couple of heuristically chosen paths. We used this insight to extend the original $1$-step method to more general estimation operators like Tree-Backup and $Q(\lambda)$. Our experiments again show how hierarchy can yield substantial performance gains over their flat counterparts, even when strictly isolating the benefit posed by credit assignment.

Finally, we conclude that hierarchy opens up an additional basis for research on return estimators. Canonically, almost all developed return estimation algorithms consider the sequential Markov Chain generated by some policy. Our hierarchical formulation of Tree-Backup and $Q(\lambda)$ shows that there exists a family of return estimation methods that also take into account the number of ways to partition traces of experiences rather than purely stepping through them.

\begin{contributions} % will be removed in pdf for initial submission,
                      % so you can already fill it to test with the
                      % ‘accepted’ class option
    Work done while J.A. de Vries was at LIACS as part of his Master thesis supervised by T.M. Moerland and A. Plaat.

\end{contributions}

\bibliography{uai2022}

\clearpage
\appendix

% The partition function and how many backup paths are theoretically plausible for estimating returns.
\section{The Number of Backup Paths} \label{ap:partition}
For multistep hierarchical methods we need to consider how to look forward (or backward) on the trace in order to compute an estimate of the value function. Whereas conventional methods do this step-by-step through time, a hierarchical backup can make time `jumps'. The backup path is thus defined by the time-deltas, suppose we have collected a trace $\tau_t$ up until time $t$ and write the time-deltas as $\Delta_\tau$ --- for flat RL this is simply $\{1\}^t_{j=0}$. The hierarchical backup needs to consider how to construct $\Delta_\tau$ such that $\sum_j (\Delta_\tau)_j = t$, this is a special case of an integer partitioning problem.

In a forward sense, the problem of computing the number of all hierarchical backup paths can also be framed as a dice rolling problem. The hierarchical environment horizon $H^a_i$ defines the maximum value that each die can take, i.e., it defines the range of the time-deltas at any level $i$. Then, the backup parameter $n$ defines the number of dice that we have. If we now assume $n$ to be sufficiently large, and write $h = H^a_i$ for shorthand, then the number of possible hierarchical backup paths of depth $t$ is given by,
\begin{align}
    \alpha_t (h, n) = \sum^n_{j=0} (-1)^j {n \choose j} { t - hj - 1 \choose n - 1}  \label{eq:backup_number}
\end{align}
where ${n \choose k}$ is the binomial coefficient.

\begin{proof}
    The quantity $\alpha_t(h,n)$ is the coefficient of the $t$-th monomial in the power expansion of rolling $n$ die with ranges $1$ to $h$, this is a result that follows from the binomial theorem (see e.g., Chapter~7 of the book by \citet{knuth1989concrete} for a similar treatment on these types of combinatorial problems). This sequence of coefficients is captured by the ordinary generating function (o.g.f.),
    \begin{align}
        F(x)    &= (x^1 + x^2 + \cdots + x^h)^n, \\
                &= x^n \left({\textstyle \sum}^{h-1}_{j=0} \: x^j \right)^{n}  \nonumber
    \end{align}
    This can be rewritten using the closed-form solution to the finite geometric series,
    \begin{align}
        F(x)    &= x^n \left( \frac{1-x^h}{1-x} \right)^n \nonumber \\
                &= x^n (1-x^h)^n(1-x)^{-n}. \nonumber
    \end{align}
    Written as a binomial series, this yields,
    \begin{align}
        F(x)    &= x^n \sum^{\infty}_{j=0} {n \choose j} (-x^h)^j  \sum^{\infty}_{k=0} {-n \choose k} (-x)^k \nonumber \\
                &= x^n \sum^{\infty}_{j=0} (-1)^j {n \choose j} x^{hj}  \sum^{\infty}_{k=0} (-1)^k {-n \choose k} x^k, \nonumber
    \end{align}
    Then see that, ${-n \choose k} = (-1)^k {k + n - 1 \choose k}$, which we can use to cancel out the other power of $-1$ (due to the even power $2k$) to give us,
    \begin{align}
        F(x)    & = x^n \sum^{\infty}_{j=0} (-1)^j {n \choose j} x^{hj}  \sum^{\infty}_{k=0} {n + k - 1 \choose k} x^k. \nonumber
    \end{align}
    To extract the coefficient of the monomial with power $t$ from this o.g.f., note that that the powers in $F(x)$ being $x^n, x^{hj}, x^k$ should together sum to $t$. In other words, $t = n + hj + k$. If we then also note that the first summation in $F(x)$ sums infinitely many zeros after $j > n$ because ${n \choose k} = 0$ for $k > n$. Then, to get the coefficient $\alpha_t$ for the $t$-th power in $F(x)$ we can omit the second summation in $F(x)$ and simply substitute for $k = t - n - hj$, which yields,
    \begin{align}
        \alpha_t x^t    &= x^n \sum^{n}_{j=0} (-1)^j {n \choose j} x^{hj}  {n+k-1 \choose k} x^{t-n-hj}, \nonumber \\
                        &= x^t \sum^{n}_{j=0} (-1)^j {n \choose j} {n+(t-n-hj)-1 \choose t-n-hj}. \nonumber
    \end{align}
    Then applying the identity ${n \choose k} = {n \choose n - k}$ and dividing the above expression by $x^t$ we get the solution in Equation~\ref{eq:backup_number}.
\end{proof}

The quantity from Equation~\ref{eq:backup_number} is of course for a fixed backup depth, because $h$ can be varied between $1, \dots, h$ there are multiple possible backup depths. When we use a smaller $n$ when computing the multistep returns, then the backup paths get truncated after $n$ steps. Then, it follows trivially that the shortest backup path is bounded by $n$ whereas the longest backup path is bounded by $hn$. We can write the total number of possible paths to compute returns on as,
\begin{align}
     S = \sum_{d=n}^{nh} \alpha_d (h, n). \label{eq:total_backup}
\end{align}
This of course yields $S= h^n$. The number of hierarchical backup paths grows at least exponentially.

With that said, we should also keep in mind that the atomic horizon $h\equiv H^a_i$ too grows exponentially with increasing levels of hierarchy according to the hierarchical action budget, $H^a_i = \prod^{i-1}_{j=0}H_j$. For example, a hierarchy with $k=3$ levels and a policy budget of $H_i=4$ for all levels $i$ gives $H^a_{k-1} = 16$. If we then want to do a $n=4$ multistep update, then in total there are approximately $\sum_{d=n}^{nh} \alpha_d(16, 4) = 16^4 \approx 65 \cdot 10^3$ possible updates. Hence, the number of backup paths actually grows super-exponentially w.r.t., $n, H_i$ and  $k$. This is not practical for any algorithm.

\subsection{Number of Tree-Backup Paths}  \label{ap:sec:tb_backup}
For our formulation of hierarchical Tree-Backup (Section \ref{sec:tb}), we provide the backup parameter $n$ and horizon $h$ when computing multistep returns. Thus, to compute all possible Tree-Backup paths, we can simply fall back to Equation~\ref{eq:total_backup}. However, this does not take into account the severe overlap between backup paths, the truncation of the traces due to the target policy probabilities, and our proposed sparsification.

In fact, our sparsification of the multistep returns are able to reduce the quantity $S$ for any $n$ and $h$ to an upper bound of just $S_{TB}=nh$ backup paths. Along with some implementation optimizations, this can be implemented in $O(n)$ time which is equivalent in time-complexity to conventional (flat) Tree-Backup (as illustrated in Algorithm~\ref{alg:HierQTB}).

% Example of how a hierarchical agent constructs an effective environment policy based on an environment trace.
\section{Example Hierarchical Policy} \label{ap:policy_example}
\begin{figure*}[p]
    \centering
    \includegraphics[width=.8\linewidth]{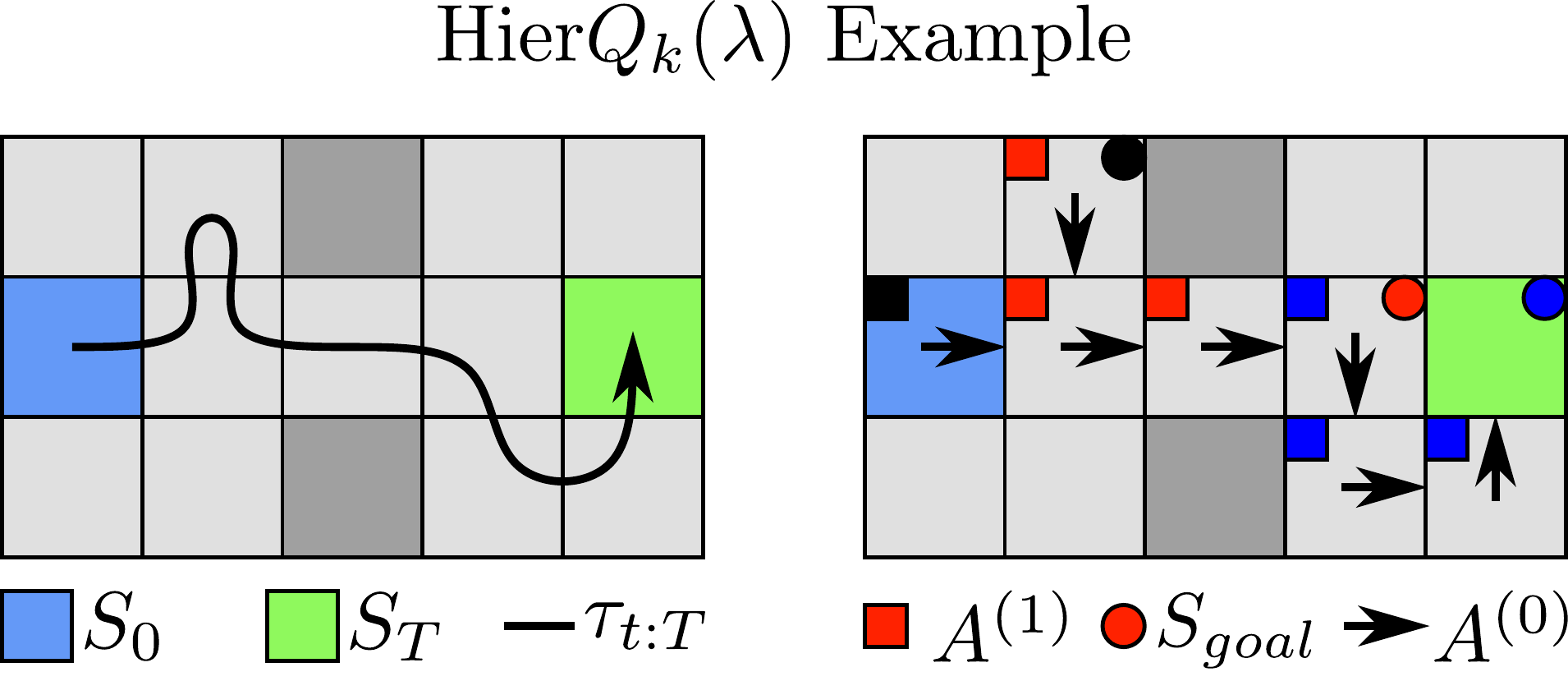}
    \caption{Greedy policy comparison of a flat (arrows) and hierarchical agent (boxes-to-circles) on an example environment trace with $\lambda=1.0$. The hierarchical agent is instructing the flat agent to take two more environment steps to reach the green tile (starting from the center in the left room) if the agent were to strictly adhere to every subgoal --- i.e., without intermediate termination. Also, the flat agent aims to follow its previously traversed path, only eliminating the small loop, whereas the hierarchical policy wants to `jump' over these redundant actions (as shown by the blue boxes-to-circle).}
    \label{fig:hierqlambdaPolicy}
\end{figure*}

To provide a better illustration to the joint effect of hierarchy on the effective environment policy, in Figure \ref{fig:hierqlambdaPolicy} we draw a $k=2$ hierarchical and $k=1$ flat policy (both conditioned on the end-goal $\langle S_T \rangle$) after one training update with a shared trace. The decomposition that the hierarchy induces was portrayed earlier in Figure~\ref{fig:hierarchy_tree}, however, that structure centered around the trace whereas Figure~\ref{fig:hierqlambdaPolicy} illustrates this in the actual environment. In the left side of Figure \ref{fig:hierqlambdaPolicy}, we see that the agent moves from the blue to the green tile in a slightly sub-optimal way, it loops before the corridor, and takes a detour before reaching $S_T$. After encountering $S_T$, the flat and hierarchical policy are both updated in parallel with backup parameters $\lambda = 1.0, \gamma = 0.95$ and $H^a_i = 3$. 

The $k=1$ flat policy exhibits a familiar pattern as indicated in the right side of Figure~\ref{fig:hierqlambdaPolicy}, its newly updated policy greedily follows along the previously traversed path (greedily by omitting the loop before the corridor). In contrast, the hierarchy seems to sample a hierarchical action that would send the lower level $i=0$ towards the loop-tile before the corridor (the one that the flat agent now omitted). Of course, this possible issue can be avoided by simply greedily re-sampling a new goal after each transition. The reason that the hierarchy sends the flat agent to this loop is actually an artifact of partitioning the hierarchical backup according to the maximal horizon $H^a_i=3$, this can be seen in the number of colored squares (left-upper aligned in each tile) that point towards the colored circles (right-upper aligned in each tile). Each hierarchical action $A^{(1)}$ is repeated on three distinct tiles. The hierarchy provides a benefit to the flat agent in that it can jump from the left-tile next to $S_T$ straight to $S_T$, without the detour that the trace made. Surely, the flat agent would still follow the detour when instructed with $S_T$, but this only shows that the hierarchical agent can look further forwards on the trace during learning. 

Thus, when the hierarchy greedily resamples goals when better ones are encountered, the hierarchical agent fundamentally supersedes the flat agent in this example. The greedy hierarchy omits the loop before the corridor, just like the flat agent. However, the hierarchy can jump over the tile left to $S_T$, straight to $S_T$ whereas the flat policy is forced to follow its previous tracks. This may indicate that it may be smart for the lower level policy to `forget' slightly in this environment, such that when $S_T$ is instructed that the agent does not greedily follow its previous trajectory. Naturally, this is closely tied to the tuning of the backup parameters $\gamma, \lambda, n$, and $H^a_i$ and the bias-variance trade off.

\begin{table*}[p]
    \centering
    
    \renewcommand{\arraystretch}{1.1}
    \caption{Table of parameter values (and ablation set) along with a short description. This table does not display every involved variable; e.g., the parameterization of the policies or the tie-breaking method in case of shared $Q$-values.}
    \label{tab:appendix:parameters}
    \renewcommand{\arraystretch}{1.1}
    \begin{tabular}{l|l|p{10cm}}
    
    Parameter & Set of Values & Description  \\
    \hline
    $\alpha$  &   $1$        &   Step-size/ learning-rate for updating value estimates ($Q$-tables) during learning (see Section~\ref{sec:background}). \\
    $H_i$  &   $3$        &   The number of \emph{hierarchical} steps before termination of the policy/ pruning this policy-branch. E.g., if policy $H_0 = 3$, then policy $\pi_0$ is terminated if it hasn't reached its goal in $3$ steps. This value was kept constant at every level $i = 0,\dots, k-1$ in the hierarchy. \\
    $b: \setS \times \setA \rightarrow [0, 1]$  &   $\{\Pi_{k-1}, \pi_0 \}$        &   Behaviour policy for generating training data. We used either the full hierarchy, or just the flat policy by fixing $A^{(i)}=S_{\text{goal}}, \forall i > 0$. \\
    $\epsilon_0$  &   $0.05, 0.25$        &   Exploration probability of the level $i=0$ policy during evaluation ($0.05$) and during training $(0.25$). \\
    $\epsilon_{i>0}$  &   $0.0$        &   Exploration probability of the level $i>0$ policies (greedy). \\
    $t_{\max}$  &   $10^5$        &   Number of allowed steps within one training episode, after which the agent is reset to the initial state. This value is kept large enough to let the agent almost always observe a reward. \\
    \hline
    \multicolumn{3}{l}{\textbf{Main Ablation Study}} \\
    \hline
     $\lambda$ &   $\{0, 0.5, 0.8, 1.0\}$  &   Decay parameter for the eligibility trace (c.f., Equation~\ref{eq:eligibility}). \\
    $n$       &   $\{1, 3, 5, 8\}$        &   Number of steps to perform Tree-Backup on for computing multistep returns (c.f., Equation~\ref{eq:TB}). \\
    $\gamma \equiv \gamma_0$  &   $0.95$        &   Base discount factor for the cumulative returns (c.f. Equation~\ref{eq:value}). \\
    $k$  &   $\{1, 2, 3, 4\}$        &   Number of hierarchy levels to define policies over policies (to sample states as goals/ actions). \\
    \hline 
    \multicolumn{3}{l}{\textbf{Backup Depth Study}} \\
    \hline
    $\lambda$ &   $1.0$  &   See Above. \\
    $n$       &   $\{9, 3, 1\}$        &   Number of adjusted Tree-Backup steps for the multistep returns such that each hierarchy level $k$ in the ablation study sent rewards back equally far (see definition of the atomic horizon $H^a_i$ in Section \ref{sec:hierq}). \\
    $\gamma$  &   $\{0.983, 0.95, 0.857\}$              &   Base discount factors adjusted proportional to the propagation depth of credits/ rewards, these were calculated as $(\gamma_0)^{1/H^a_{k-1}}$ (see Above). \\
    $k$       &   $\{1, 2, 3\}$        &   See Above. \\
                        
\end{tabular}
\end{table*}

% Note: Figure below illustrates how a hierarchical Q-learning policy would behave based on the given environment trace for training. This is meant to give intuition on how the effective environment policy is constructed and how the higher level policy instructs the lower level policy.

% Algorithmic ablations (e.g., greedy options), hyperparameter settings (epsilon), etc.
\section{Further Experiment Details} \label{ap:experiment}
This section enumerates all further experimental details that did not fit in the main narrative. For an overview of all hyperparameters and the domain we considered for our ablations, see Table~\ref{tab:appendix:parameters}.

For all our main experiments we utilized a greedy hierarchy during training $\epsilon = 0$, but we also greedily terminated and resampled goals for the lower levels to follow \emph{during evaluation time}. So sub-optimal goals could be sampled and followed during training, but were resampled greedily according to $Q_i$ during test-time. Furthermore, for all eligibility trace based agents, we truncated/ cut the traces when the eligibility value would shrink to $z < 10^{-8}$ to reduce computation time (see Appendix~\ref{ap:heuristics} for further reference on our implementation). We also did not allow hierarchical policies to sample a state as goal that they were already positioned at $\pi_i(S_t | S_t, \textbf{g}) = 0, \forall i > 0$. If the instructed goal state of a hierarchical policy $i$ was within reach of the agent, than that policy would directly sample the goal (if the agent sees the goal, then it can greedily sample the goal). We utilized uniformly random tie breaking in case of shared $Q$ values for various state-action pairs, at every level $i=0, \dots, k-1$.

All experiments were run on an AMD Ryzen 9 5900x CPU which took approximately a 1-2 days of compute when running multiple ablations in parallel. Our framework was built on Python 3.9.7 with Numpy 1.20.3, OpenAI Gym 0.19.0, and we utilized environment examples from the public MazeLab library by \citet{mazelab}. Code for implementation and reproduction of our experiments is available at: <LINK REMOVED FOR DOUBLE-BLIND>, see also the supplementary material for the experiment data used for visualization.

%Replace link on acceptance: \href{https://github.com/joeryjoery/HierQ/tree/UAI2022}{\url{https://github.com/joeryjoery/HierQ/tree/UAI2022}}

% Notes on the original method Vs. ours. For example: sparse binary Vs. penalizing rewards for the goal-specialized policies. Also the optimizations for the action-spaces: relative action spaces through spatial neighborhoods compared to the absolute state-space. Also: the top-level in the hierarchy need not estimate |S| policies as it is always conditioned on the environment reward. Etc...
\section{Implementation Notes} \label{ap:heuristics}
This section discusses implementation specific details and considerations. The next subsection describes important points about our hierarchical $Q$-learning algorithm and HierTB$_k(n)$. The subsections afterward discuss our policy representation and choice for reward function. We provide an additional experimental result/ ablation there to illustrate and back our choice for these particular implementations.

\subsection{Algorithm Details}
% Semi-backward view.
In Algorithm~\ref{alg:HierQLambda} one may have noticed that we can send the rewards back in a form of `semi-backward' view in order to update $1$-step state-action pairs (line 9). It turns out that for all our hierarchical return operators, we don't need to wait for the entire set $\Delta_t^{(i)}$ to be computed before being able to update $Q$-estimates. In fact, we can perform the hierarchical update by only computing $\vecG^1_t$ (Equation~\ref{eq:hierQ}) and updating the $Q$-estimates towards this return value for all state-action pairs in $\{S_{t-j + 1}\}_{j=1}^{H^a_i} \times \{A_t^{(i)}\}$. This mechanism is `semi-backward' in the sense that $\textbf{S}^{(i)}_t = \{S_{t-j + 1}\}_{j=1}^{H^a_i}$ contains all \emph{past} states for which $A_t^{(i)}$ is a reachable action. The motivation behind this particular mechanism can also be seen by noting that $\vecG^1_t \equiv \vecG^j_{t-j}, \forall j > 0$; there is exact overlap between returns within a rolling window of states.

% Tree-Backup Further explanation:
We also adopted this efficient semi-backward view for our Hierarchical Tree-Backup implementation, which is shown in Algorithm~\ref{alg:HierQTB}. The algorithm is of almost the exact same form as conventional Tree-Backup \citep{sutton_reinforcement_2018}, but differs in the fact that: at each backup-step we perform `jumps' of length $H^a_i$ in time, and given the return $\textbf{G}$ we update multiple state-action pairs (inferred from the trailing states $\textbf{S}^{(i)}_{t_n}$.

Like conventional Tree-Backup, our algorithm incurs a time-delay proportional to $n$ before being able to compute update targets. This is exacerbated by the hierarchy horizon $H^a_i$, we need to wait $nH^a_i$ time-steps before being able to compute the multistep return. Despite this, we can still compute returns/ update targets at each timestep due to the observed pseudo-rewards. This allows us to continually update all goal-specialized levels $\Pi_i, i = 0, \dots, k-1$. As mentioned before and as illustrated by Algorithm~\ref{alg:HierQTB}, due to our sparsification assumption and the overlap between hierarchical returns at different timesteps, $\vecG_t^1 \equiv \vecG_{t-j}^j$ (Equation~\ref{eq:hierQ}), we only have to backup along a single path to update the $Q$-values of the elements in $\textbf{S}^{(i)}_t \times \{A_t^{(i)}\}$.

\begin{algorithm}[tb] 
    \DontPrintSemicolon
	\caption{Tree-Backup for Hierarchical $Q$-learning.}
	\label{alg:HierQTB}
	\nonl \textbf{Input:} Environment trace $\tau_{t+1} = \{S_j, A_j\}_{j=0}^{t+1}$, the hierarchy level $i$, a target policy $\pi$, and $Q$-table $Q_i(s, a, g)$  \;
	\nonl \textbf{Parameter:} Discount factor $\gamma \in [0, 1)$, step-size $\alpha \in (0, 1]$, backup depth $n \in \mathbb{N}$, and policy reach $H^a$ \;
	\BlankLine
    $t_n \gets t - H^a(n - 1)$  \hspace{\fill} \{Sweep $n$ to $1$ if $t=T$ \} \;
    \If{$t_n \ge 0$}{
        $\textbf{G} \gets \textbf{r}_{t+1} + \bm{\gamma}_{t+1}  \mathbb{E}_\pi Q_i(S_{t+1}, \cdot, \textbf{g})$ \;
        \For{$k = t - H^a + 1 \ge t_n$ with steps of $-H^a$} {
            $A \gets S_{k+H^a}$ \textbf{if} $i > 0$ \textbf{else} $A \gets A_k$ \;
            $\textbf{G} \gets \textbf{r}_k + \bm{\gamma}_k (\bm{\pi}_k \textbf{G} + (1 - \bm{\pi}_k) \mathbb{E}_\pi Q_i(S_{k}, \cdot, \textbf{g}))$ \;
        }
        $A \gets S_{1 + t_n}$ \textbf{if} $i > 0$ \textbf{else} $A \gets A_{t_n}$ \;
        \For{$j = 0, 1, \dots, \min(H^a - 1, t_n)$}{
            $Q_i(S_{t_n-j}, A, \textbf{g}) \gets (1-\alpha)Q_i(S_{t_n-j}, A, \textbf{g}) + \alpha \textbf{G}$ \;
        }
    }
\end{algorithm}

% IJCAI FORMAT
% \begin{algorithm}[tb] 
% 	\caption{Tree-Backup for Hierarchical $Q$-learning.}
% 	\label{alg:HierQTB}
% 	\textbf{Input:} Environment trace $\tau_{t+1} = \{S_0, A_0, S_1, A_1, \dots\}$, a target policy $\pi$, hierarchy level $i$ and $Q$-table $Q_i(s, a, g)$. \\
% 	\textbf{Parameter:} Discount factor $\gamma \in [0, 1)$, step-size $\alpha \in (0, 1]$, backup depth $n \in \mathbb{N}$, and effective policy horizon $H^a$ \\
% 	\begin{algorithmic}[1]
% 	    \STATE $t_n \gets t - H^a(n - 1)$  \hspace{\fill} \COMMENT{Sweep $n$ to $1$ if $t=T$}
% 	    \IF{$t_n \ge 0$}
% 	        \STATE $\textbf{G} \gets \textbf{r}_{t+1} + \bm{\gamma}_{t+1}  \mathbb{E}_\pi Q_i(S_{t+1}, \cdot, \textbf{g})$
% 	        \FOR{$k = t - H^a + 1 \ge t_n$ with steps of $-H^a$}
% 	            \STATE Infer $A \gets S_{k+H^a}$ \textbf{if} $i > 0$ \textbf{else} $A \gets A_k$
%                 \STATE $\textbf{G} \gets \textbf{r}_k + \bm{\gamma}_k (\bm{\pi}_k \textbf{G} + (1 - \bm{\pi}_k) \mathbb{E}_\pi Q_i(S_{k}, \cdot, \textbf{g}))$
%             \ENDFOR
%             \STATE Infer $A \gets S_{1 + t_n}$ \textbf{if} $i > 0$ \textbf{else} $A \gets A_{t_n}$
%             \FOR{$j = 0, 1, \dots, \min(H^a - 1, t_n)$}
%                 \STATE $Q_i(S_{t_n-j}, A, \textbf{g}) \gets (1-\alpha)Q_i(S_{t_n-j}, A, \textbf{g}) + \alpha \textbf{G}$
%             \ENDFOR
%         \ENDIF
% 	\end{algorithmic}
% \end{algorithm} 

% HierQ(Lambda) Further Explanation:
\subsubsection*{Eligibility Traces}
At a first glance, it might seem that an eligibility trace would make the extension for $1$-step hierarchical updates to multistep methods even simpler opposed to Tree-Backup. With eligibility traces we can just update recency values within the trace and cast a $1$-step error backwards (see Equation~\ref{eq:qlambda}). However, the hierarchical structure actually makes the extension to Tree-Backup$(\lambda)$ quite intricate.

As alluded to in the main paper (Section~\ref{sec:hql}), there are numerous subtleties in extending our version of Tree-Backup to a full backward method. Since every level in the hierarchy estimates the returns at different temporal resolutions, this means that we need to track $k$ separate eligibilities --- in fact, any level $i$ can be interpreted to estimate the return up to a $(\gamma)^{1/H^a_i}$ diluted discount (and/ or decay rate). Moreover, in order to correctly update all policies $\pi \in \Pi_i$, at every level $i$, we need to account for the policy correction terms at each transition for each goal-specialized policy. For example, an action $A_t$ can be optimal at $S_t$ for reaching goal $S'$ but be suboptimal for reaching another goal $S''$. Hence, why we argued for utilizing $|\setS|$ separate eligibility traces at each separate level $i$.

Another interesting implication was that we had to use another $H^a_i$ separate eligibility traces for each temporal horizon. This is an artifact of our heuristic choice of dealing with the exorbitant number of possible backup paths by making fixed time-jumps of $H^a_i$ steps. Suppose we applied the update in Equation~\ref{eq:hiertrace} at every environment step, then it is easy to see that our sparse structure gets violated: we do not only add new paths to the previously added ones, but we also decay all previous path exponentially fast due to the repeated multiplication with $(\gamma \lambda)$. For example, if we added a state-action pair of time-length $H^a_i$ in the previous time-step, and in the next time-step we added the state-action pairs for all trailing states, we have a backup-path that connects through a $H^a_i$-time jump to an intermediate $1$-step path. From Algorithm~\ref{alg:HierQTB} it should be obvious that intermediate time-jumps should always be $H^a_i$.

\subsection{Policy Representation and Memory}
As explained in Section~\ref{sec:hierq} each level within the hierarchy was parameterized as $\setA_i \subseteq \Pi_{i-1}, i>0$. The action space of the hierarchical levels was the set of goal-specialized polices at the level below. This is quite trivial to implement simply by using the \emph{full} set of polices $\setA_i \equiv \Pi_{i-1}, i > 0$ such that the agent can figure out itself which goals are reachable (which actions are viable) through the backed up rewards. Seeing as we utilized fully greedy hierarchical policies, after observing the first reward, the agent would always sample valid state-actions. 

We restricted this set of policies by defining $\setA$ to be a function over the power set of the state-space (in terms of goal-policies), $\setA_i: \setS \rightarrow \mathcal{P}(\Pi_{i-1})$ such that each $\pi \in \setA_i(S_t)$ could actually reach its goal within $H^a_i$ steps. In other words, we restricted the action set of each hierarchical policy at every state such that any sampled policy's goal state lied within a $H^a_i$ neighborhood of the current state in terms of the state-space (we utilized $l_1$ neighborhoods). This makes sense because any policy that the hierarchical level could sample that lied outside this neighborhood would be terminated prematurely due to the hierarchical action budgets $H_i$. Of course, this is an \emph{ad-hoc} detail, and is not applicable in every domain; this either requires knowledge of the transition function or an assumption on the environment's geometry. Due to our tabular domain, this assumption was necessary to cut down on the memory usage. 

Another memory optimization is noting that we do not utilize \emph{all} policies in the top-level $\Pi_{k-1}$, only the policy conditioned on the environment reward. This cut down the memory complexity from $O(|\setS|^2 \times |\setA_{k-1}|)$ to $O(|\setS| \times |\setA_{k-1}|)$ for the top-level. Combined with the restricted action-spaces at the lower levels $i < k-1$, this provided a tractable implementation for the larger domains (e.g., $20 \times 20$ Gridworld).

\subsection{Reward Function and Discounting}
A common theme in our main paper was whether hierarchy simply adds depth to the reward estimation similarly to flat multistep methods. However, we argued that hierarchy creates skip-connections over time whereas conventional methods simply step-through the trace. An important factor that balanced the depth of the return in either case, was the discount factor $\gamma$. Due to our choice of binary pseudo-rewards, i.e., the Successor Representation \citep{dayan_improving_1993}, after sufficient time the rewards will get diluted due to the exponential discounting. For our experiments/ ablations it may have made sense to utilize another reward function, such that $\gamma$ could be set to $\gamma=1$. A simple alternative is $\textbf{r}_t' = 1 - \textbf{r}_t$, i.e., a penalizing reward of $-1$ at each time-step and $0$ when the goal is observed.

Though this choice for reward function eliminates one confounder $\gamma$, it introduces another one. The penalizing reward has a similar effect to count-based exploration methods. During training episodes, states will get penalized continuously until goals are observed, this has the effect that the agent will steer away from frequently visited states if it doesn't observe rewards for said goals. Our choice of the binary rewards ensured that the agents would stay uniformly random until actual rewards were observed.

\subsection{Policy and Reward Ablation}
\begin{figure*}[p]
    \centering
    \includegraphics[width=.9\linewidth]{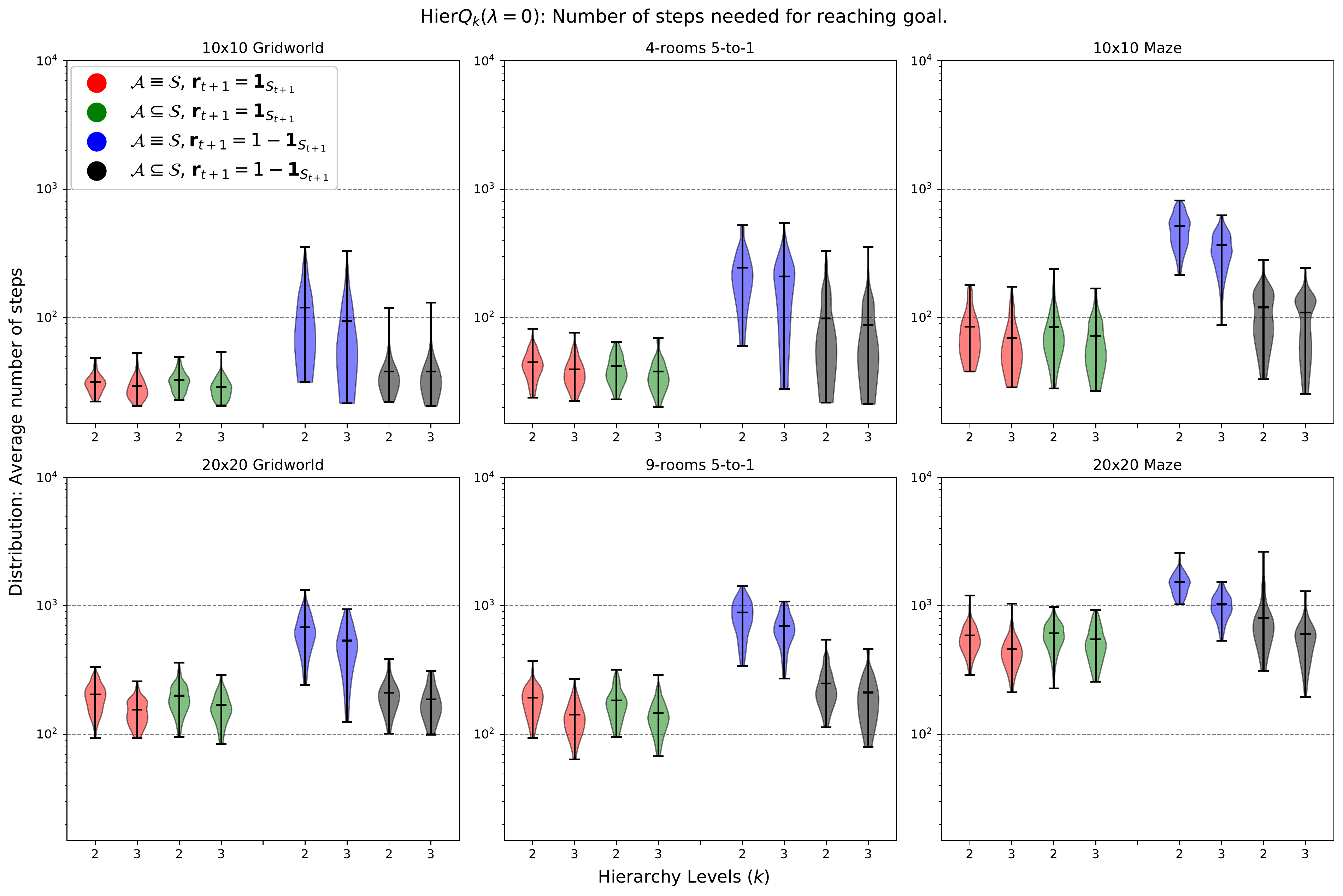}
    \caption{Comparison of our implementation for Hier$Q_k(\lambda = 0)$ to that of \citet{levy_learning_2018}. The $x$-axis shows the number of hierarchy levels $k$, split by the reward function and as indicated in the legend. The $y$-axis shows the aggregate number of steps to reach the environment goal of the greedy policy over the first $50$ training episodes, the violinplots illustrate their distribution over $200$ random seeds.}
    \label{fig:binary_rewards}
\end{figure*}

Our choice for the relative policy parameterization and binary pseudo-rewards is motivated by our additional ablation results in Figure~\ref{fig:binary_rewards}. These results were generated with the \emph{exact} same experimental setup as discussed before in Section~\ref{sec:experiments}. However, here we performed ablations over the choice of reward function (binary Vs. penalizing) and the policy parameterization (restricted Vs. unrestricted) only for the $1$-step Hierarchical $Q$-learning algorithm. In a sense, this can be seen as a direct comparison to the original algorithm by \citet{levy_learning_2018} and our adaptation. All-in-all, these results simply strengthen our implementation considerations that we used to generate our main results.

% Note: Figure below illustrates the ablation performance scores (same as in the main paper) when generating data for training using a flat policy (fixed hierarchy with all intermediate goals = end-goal). This figure is slightly more difficult to read (more overlap between lines) and is aimed to eliminate the confounding variable of the hierarchical training for comparing performance scores. This means we can get a directer comparison on the effect of 'backup depth' in agent performance. 
\begin{figure*}[p]
    \centering
    \includegraphics[width=\linewidth]{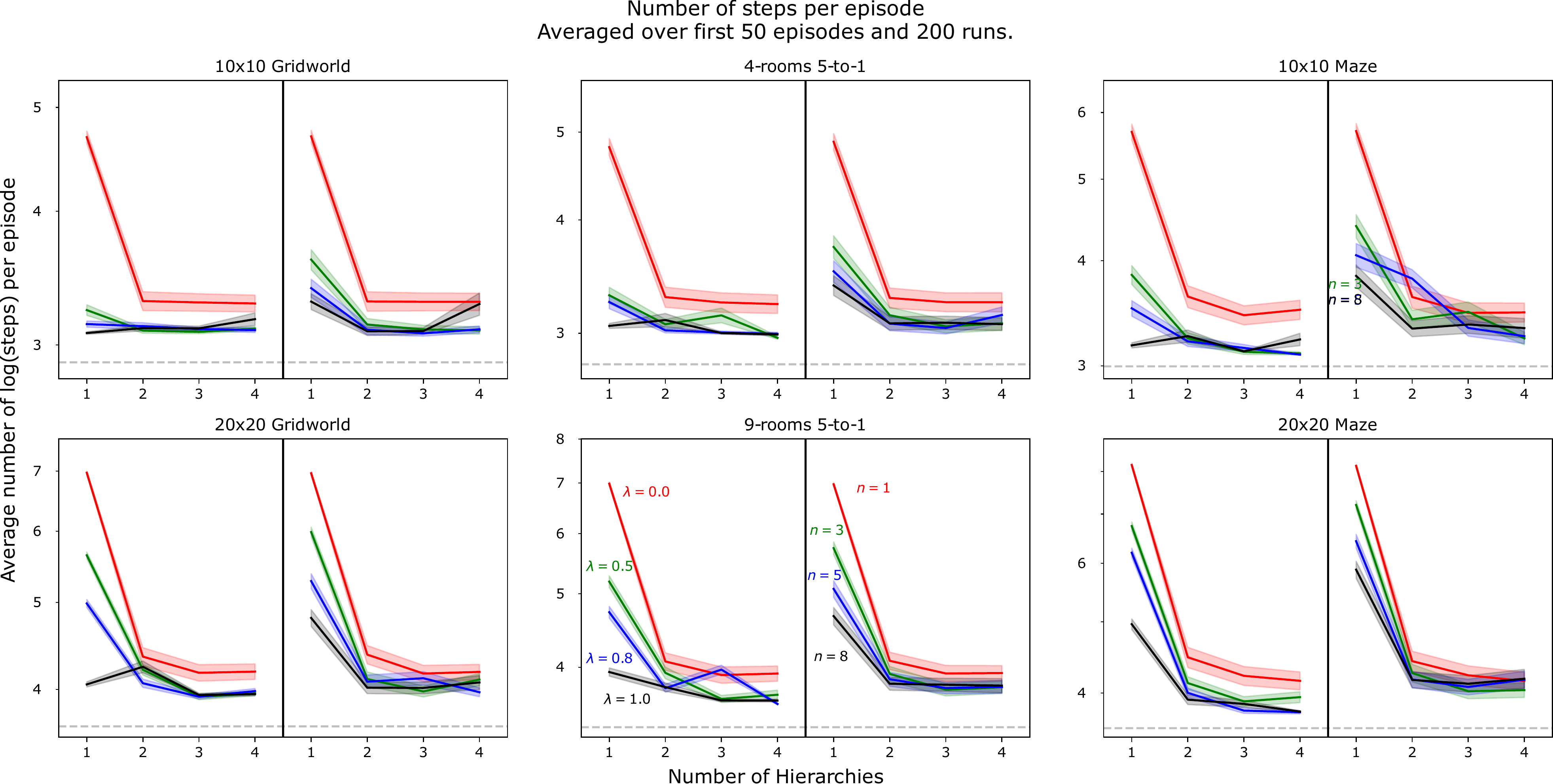}
    \caption{Marginal log-performance of each experiment configuration for each environment from Figure~\ref{fig:ablations_envs} (lower is better). This figure is identical to the main paper (including the $y$-scale; c.f., Figure~\ref{fig:ablations_mean}), however this data portrays the agent's performance when training proceeded purely with the flat policy (i.e., by keeping the hierarchy fixed during training: $A^{(i)} = S_{\text{goal}}, \forall i > 0$) and evaluation with the hierarchical policy.}
    \label{fig:flat_ablations_mean}
\end{figure*}

% Note: Figure below is the violin-plot comparing the absolute Vs. relative action space for the policy parameterization, and the binary sparse reward Vs. the -1 rewards. The reward comparison is interesting as the discount factor \gamma can be set to 1 for this reward function (not for the binary). Ultimately eliminating a confounding variable for our analyses of the reward propagation depth.

\begin{table*}[p]
    \centering
    \caption{Table of all first episode reward statistics from the main ablation study, i.e., the recorded number of environment steps in the first training episode. As in the main paper (c.f., Table \ref{tab:first_reward}), the cell values contain the mean and standard errors $\mu \pm s^{-1}$ rounded to integers. These means can also be interpreted as the scale parameter $\beta$ for an exponential distribution Exp$(\beta^{-1})$, though this data is best captured by a scaled Poisson (or Gamma). The lowest means are emphasized with bold fonts whereas the highest means are emphasized with italics, see also the supplementary material for further visualizations. }
    \label{tab:appendix:first_reward}
    \renewcommand{\arraystretch}{1.1}
\begin{tabular}{p{1.2cm}p{1cm}|p{2cm}p{2cm}p{2cm}p{2cm}p{2cm}p{2cm}}
\multicolumn{2}{c|}{Hier$Q_k(\lambda)$} &  10x10 \newline Gridworld & 20x20 \newline Gridworld & 4-rooms \newline 5-to-1  & 9-rooms \newline 5-to-1 & 10x10 Maze & 20x20 Maze  \\
\hline
$\lambda=0$         & $k=1$ &  $\bm{561 \pm 37}$        &  $\bm{2746 \pm 178}$      &  $\bm{931 \pm 56}$        &  $\bm{2556 \pm 142}$      &  $2149 \pm 173$           &  $9215 \pm 599$           \\
                    & $k=2$ &  $834 \pm 63$             &  $5188 \pm 380$           &  $1135 \pm 66$            &  $3949 \pm 230$           &  $\mathit{5418 \pm 734}$  &  $\mathit{14312 \pm 982}$ \\
                    & $k=3$ &  $857 \pm 64$             &  $\mathit{6996 \pm 529}$  &  $\mathit{1293 \pm 69}$   &  $\mathit{4178 \pm 277}$  &  $2497 \pm 209$           &  $12388 \pm 791$          \\
                    & $k=4$ &  $\mathit{950 \pm 71}$    &  $6704 \pm 548$           &  $1200 \pm 70$            &  $4035 \pm 238$           &  $\bm{1471 \pm 127}$      &  $\bm{10665 \pm 688}$     \\
\hline
$\lambda=0.5$       & $k=1$ &  $\bm{626 \pm 35}$        &  $\bm{3168 \pm 179}$      &  $\bm{1024 \pm 59}$       &  $\bm{2541 \pm 125}$      &  $2797 \pm 215$           &  $\bm{9080 \pm 629}$      \\
                    & $k=2$ &  $1043 \pm 83$            &  $6014 \pm 478$           &  $1222 \pm 74$            &  $\mathit{4136 \pm 239}$  &  $\mathit{4805 \pm 564}$  &  $\mathit{14028 \pm 908}$ \\
                    & $k=3$ &  $990 \pm 70$             &  $8244 \pm 590$           &  $1273 \pm 91$            &  $4109 \pm 283$           &  $2860 \pm 262$           &  $11647 \pm 784$          \\
                    & $k=4$ &  $\mathit{1245 \pm 90}$   &  $\mathit{8770 \pm 727}$  &  $\mathit{1339 \pm 77}$   &  $3928 \pm 270$           &  $\bm{1308 \pm 95}$       &  $10580 \pm 643$          \\
\hline
$\lambda=0.8$       & $k=1$ &  $\bm{629 \pm 36}$        &  $\bm{3300 \pm 189}$      &  $\bm{941 \pm 54}$        &  $\bm{2792 \pm 151}$      &  $2525 \pm 187$           &  $10384 \pm 684$          \\
                    & $k=2$ &  $1130 \pm 78$            &  $6528 \pm 436$           &  $1308 \pm 73$            &  $3717 \pm 213$           &  $\mathit{6318 \pm 665}$  &  $13167 \pm 867$          \\
                    & $k=3$ &  $1134 \pm 76$            &  $8254 \pm 663$           &  $\mathit{1400 \pm 99}$   &  $\mathit{4655 \pm 303}$  &  $2914 \pm 267$           &  $\mathit{14865 \pm 944}$ \\
                    & $k=4$ &  $\mathit{1315 \pm 105}$  &  $\mathit{8406 \pm 788}$  &  $1331 \pm 71$            &  $4492 \pm 287$           &  $\bm{1663 \pm 143}$      &  $\bm{9854 \pm 633}$      \\
\hline
$\lambda=1$         & $k=1$ &  $\bm{562 \pm 31}$        &  $\bm{3023 \pm 216}$      &  $\bm{838 \pm 46}$        &  $\bm{2937 \pm 162}$      &  $2236 \pm 189$           &  $\bm{8922 \pm 574}$      \\
                    & $k=2$ &  $1069 \pm 76$            &  $6586 \pm 504$           &  $1255 \pm 72$            &  $3532 \pm 198$           &  $\mathit{5617 \pm 600}$  &  $\mathit{13826 \pm 829}$ \\
                    & $k=3$ &  $1010 \pm 70$            &  $\mathit{10385 \pm 819}$ &  $1339 \pm 76$            &  $\mathit{4102 \pm 240}$  &  $2965 \pm 260$           &  $11590 \pm 711$          \\
                    & $k=4$ &  $\mathit{1380 \pm 104}$  &  $8643 \pm 766$           &  $\mathit{1370 \pm 81}$   &  $3933 \pm 235$           &  $\bm{1785 \pm 151}$      &  $10923 \pm 707$          \\
\hline
& & & & & & & \\
\multicolumn{2}{c|}{Hier$\text{TB}_k(n)$} & & & & & & \\
\hline
$n=1$               & $k=1$ &  $\bm{637 \pm 43}$        &  $\bm{3251 \pm 221}$      &  $\bm{895 \pm 52}$        &  $\bm{2960 \pm 190}$      &  $2255 \pm 155$           &   $\bm{ 9718 \pm 651}$    \\
                    & $k=2$ &  $859 \pm 70$             &  $5518 \pm 368$           &  $1196 \pm 80$            &  $3568 \pm 219$           &  $\mathit{4464 \pm 614}$  &  $\mathit{14190 \pm 1040}$\\
                    & $k=3$ &  $\mathit{959 \pm 71}$    &  $6417 \pm 526$           &  $\mathit{1297 \pm 69}$   &  $\mathit{4658 \pm 279}$  &  $2469 \pm 199$           &  $12969 \pm 832$          \\
                    & $k=4$ &  $895 \pm 72$             &  $\mathit{6468 \pm 605}$  &  $1251 \pm 74$            &  $4244 \pm 267$           &  $\bm{1490 \pm 105 }$       &  $9919 \pm 585$         \\
    \hline
$n=3$               & $k=1$ &  $\bm{698 \pm 44}$        &  $\bm{3064 \pm 210}$      &  $\bm{972 \pm 58}$        &  $\bm{3183 \pm 193}$      &  $2493 \pm 171$           &   $\bm{ 9028 \pm 555}$    \\
                    & $k=2$ &  $1069 \pm 87$            &  $5176 \pm 394$           &  $1174 \pm 75$            &  $3627 \pm 206$           &  $\mathit{5000 \pm 555}$  &  $\mathit{12388 \pm 936}$ \\
                    & $k=3$ &  $994 \pm 76$             &  $\mathit{8010 \pm 600}$  &  $\mathit{1310 \pm 77}$   &  $\mathit{4776 \pm 332}$  &  $2573 \pm 247$           &  $11168 \pm 748$          \\
                    & $k=4$ &  $\mathit{1163 \pm 78}$   &  $6563 \pm 628$           &  $1262 \pm 71$            &  $3880 \pm 279$           &  $\bm{1681 \pm 143}$      &  $9549 \pm 584$           \\
\hline
$n=5$               & $k=1$ &  $\bm{519 \pm 33}$        &  $\bm{3023 \pm 186}$      &  $\bm{946 \pm 55}$        &  $\bm{2873 \pm 154}$      &  $2514 \pm 185$           &  $\bm{9440 \pm 644}$      \\
                    & $k=2$ &  $1050 \pm 86$            &  $6077 \pm 480$           &  $1090 \pm 63$            &  $3723 \pm 219$           &  $\mathit{4097 \pm 420}$  &  $\mathit{13469 \pm 884}$ \\
                    & $k=3$ &  $869 \pm 69$             &  $8179 \pm 649$           &  $1234 \pm 78$            &  $\mathit{4589 \pm 385}$  &  $2824 \pm 251$           &  $10385 \pm 637$          \\
                    & $k=4$ &  $\mathit{1141 \pm 91}$   &  $\mathit{8313 \pm 731}$  &  $\mathit{1292 \pm 78}$   &  $4236 \pm 236$           &  $\bm{1599 \pm 146}$      &  $9846 \pm 634$           \\
\hline
$n=8$               & $k=1$ &  $\bm{632 \pm 36}$        &  $\bm{3139 \pm 191}$      &  $\bm{1064 \pm 61}$       &  $\bm{2789 \pm 161}$      &  $2479 \pm 193$           &  $\bm{8984 \pm 597}$      \\
                    & $k=2$ &  $968 \pm 74$             &  $6094 \pm 404$           &  $1131 \pm 68$            &  $3591 \pm 218$           &  $\mathit{3707 \pm 439}$  &  $\mathit{14032 \pm 933}$ \\
                    & $k=3$ &  $1030 \pm 74$            &  $7691 \pm 557$           &  $1271 \pm 74$            &  $3891 \pm 233$           &  $2900 \pm 256$           &  $11853 \pm 753$          \\
                    & $k=4$ &  $\mathit{1140 \pm 80}$   &  $\mathit{8691 \pm 830}$  &  $\mathit{1341 \pm 78}$   &  $\mathit{4310 \pm 243}$  &  $\bm{1551 \pm 146}$      &  $9077 \pm 559$           \\
\end{tabular}
\end{table*}

% Note: Figure below illustrates the ablation loss curves (same as in the main paper) when appropriately balancing credit assignment. This figure contains the complete ablation and is slightly more difficult to read (more overlap between lines). The adjusted credit assignment parameters shows how performance is influenced by the hierarchy.
\begin{figure*}[p]
    \centering
    \includegraphics[width=\linewidth]{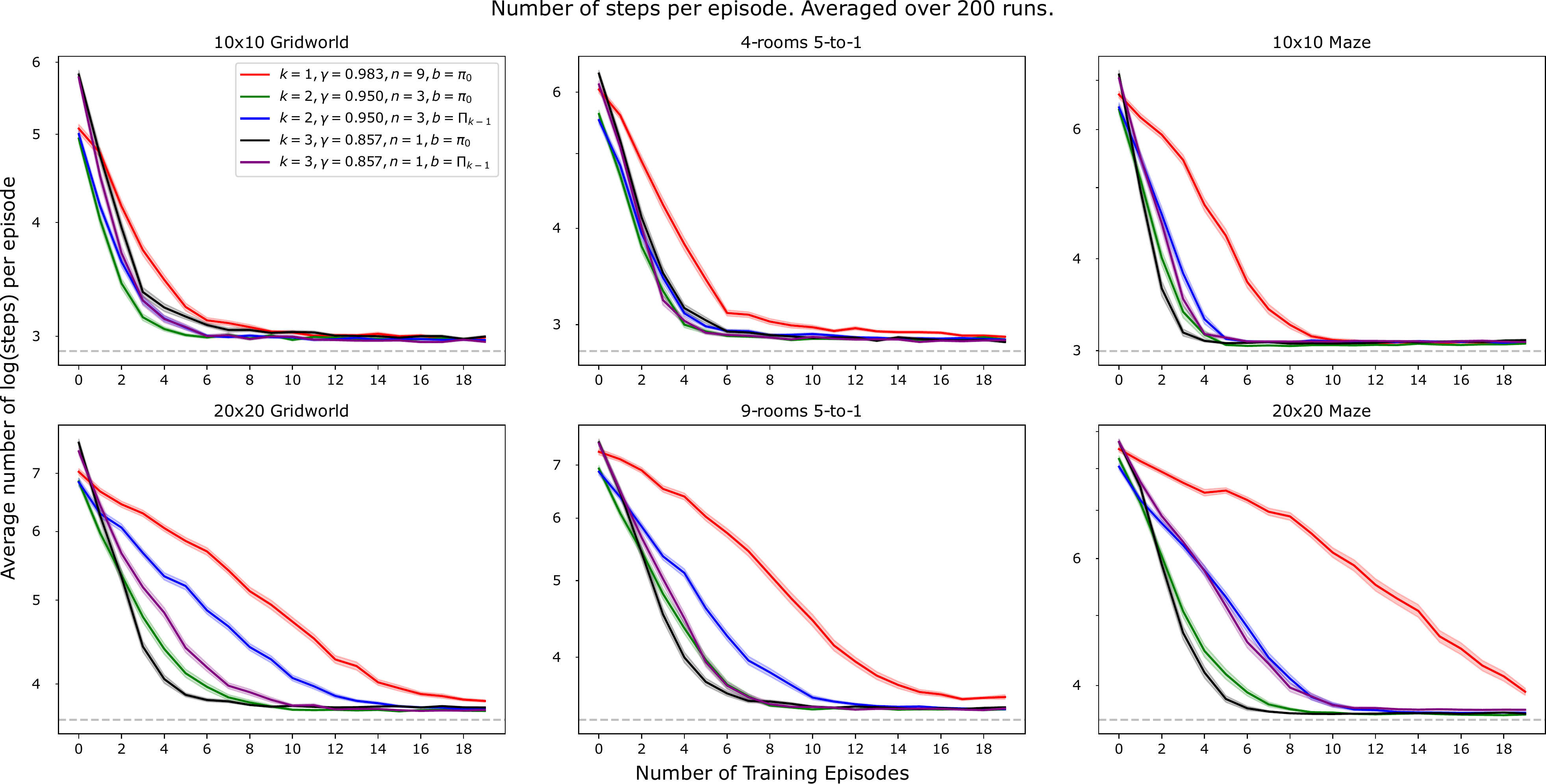}
    \caption{Average $\log$(score) of the evaluated hierarchical Tree-Backup agents at each training episode (lower is better) when credit assignment \emph{depth} is appropriately balanced for each hierarchy level $k$. Shaded regions indicate $1$-standard error of the mean.}
    \label{fig:full_loss_curve_n}
    
    \bigskip
    
    \includegraphics[width=\linewidth]{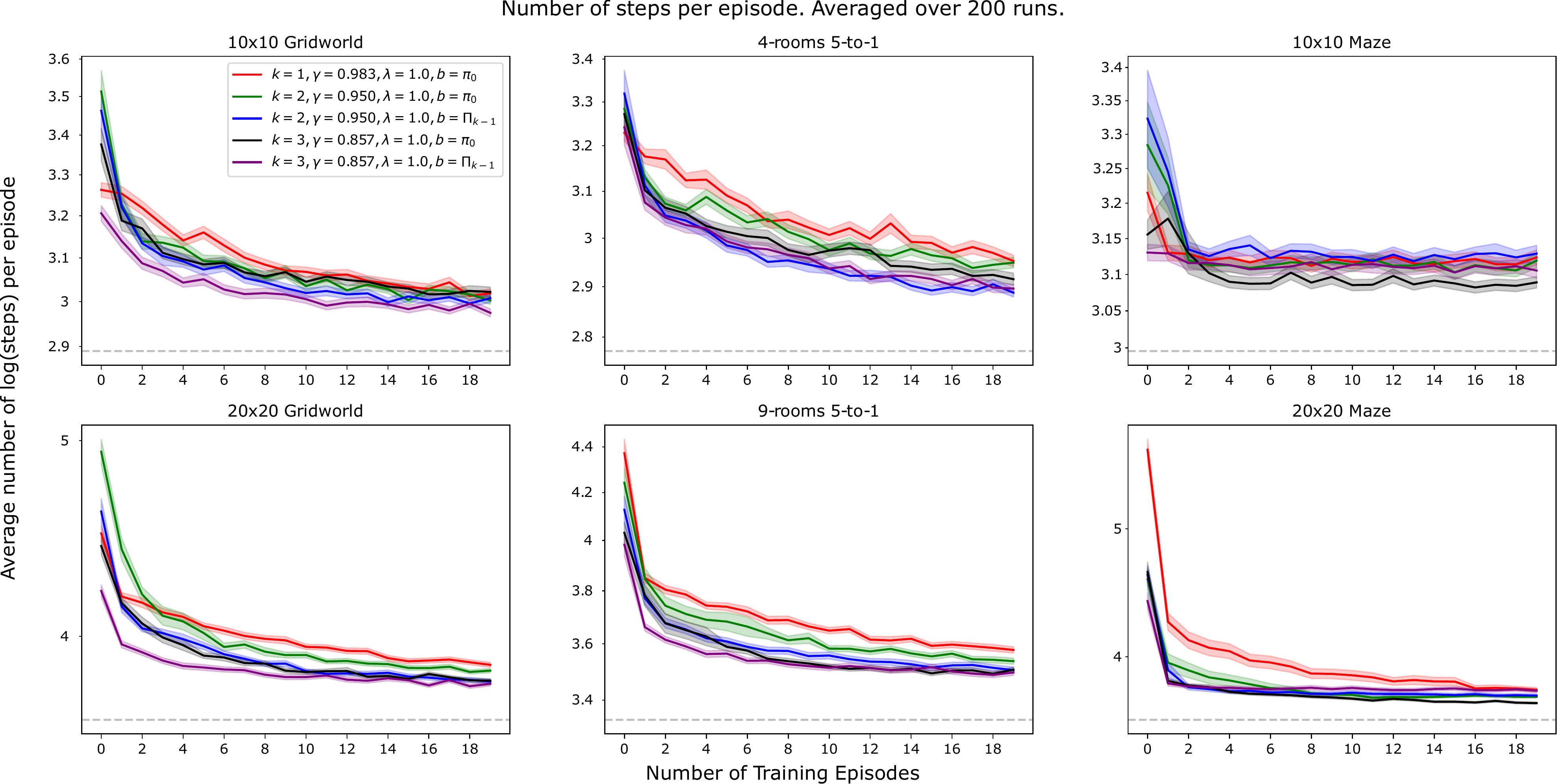}
    \caption{Average $\log$(score) of the evaluated hierarchical $Q(\lambda)$ agents at each training episode (lower is better) when credit assignment \emph{depth} is appropriately balanced for each hierarchy level $k$. Shaded regions indicate $1$-standard error of the mean.}
    \label{fig:full_loss_curve_lambda}
\end{figure*}

\end{document}